\documentclass{article}

\usepackage[preprint]{neurips_2025}

\usepackage[utf8]{inputenc} % allow utf-8 input
\usepackage[T1]{fontenc}    % use 8-bit T1 fonts
\usepackage{hyperref}       % hyperlinks
\usepackage{url}            % simple URL typesetting
\usepackage{booktabs}       % professional-quality tables
\usepackage{arydshln}
\usepackage{amsfonts}       % blackboard math symbols
\usepackage{nicefrac}       % compact symbols for 1/2, etc.
\usepackage{microtype}      % microtypography
\usepackage{wrapfig}
\usepackage{graphicx}
\usepackage{fontawesome}
\usepackage{multirow}
\usepackage{amsmath}
\usepackage{siunitx}
\usepackage[font=footnotesize]{caption}
\usepackage{makecell}
\usepackage[table,xcdraw]{xcolor}
\usepackage[linesnumbered,ruled,vlined]{algorithm2e}
\usepackage{pifont}

\newcommand{\etc}{\textit{etc}}  %
\newcommand{\cg}{\color{gray}}
\newcommand{\pos}{\ \ \ -\ \ \ \ }
\definecolor{lightgray}{rgb}{0.90, 0.90, 0.90}
\definecolor{lightblue}{rgb}{0.80, 0.90, 0.95}
\definecolor{lighterblue}{rgb}{0.88, 0.94, 0.98}

\usepackage{paralist}

\title{Threading \textit{Keyframe} with \textit{Narratives}: \\
MLLMs as Strong Long Video Comprehenders}

\author{
Bo Fang$^{1}$,\ Wenhao Wu$^{2,3}$,\ Qiangqiang Wu$^1$,\ Yuxin Song$^2$,\ Antoni B. Chan$^{1}$\thanks{Corresponding author} \\
$^1$Department of Computer Science, City University of Hong Kong \\
$^2$Baidu Inc. \quad $^3$University of Sydney \\
\texttt{\{bofang6-c,qiangqwu2-c\}@my.cityu.edu.hk, wenhao.wu@sydney.edu.au}\\
\texttt{songyuxin02@baidu.com, abchan@cityu.edu.hk}
}

\begin{document}

\maketitle

\begin{abstract}
  Employing Multimodal Large Language Models (MLLMs) for long video understanding remains a challenging problem due to the dilemma between the substantial number of video frames (i.e., visual tokens) versus the limited context length of language models.
  Traditional uniform sampling often leads to selection of irrelevant content,  while post-training MLLMs on thousands of frames imposes a substantial computational burden.
  In this paper, we propose \textit{threading keyframes with narratives} (\textbf{Nar-KFC}), a plug-and-play module to facilitate effective and efficient long video perception. 
  Nar-KFC generally involves two collaborative steps.
  First, we formulate the \textit{keyframe} selection process as an integer quadratic programming problem, jointly optimizing query-relevance and frame-diversity. To avoid its computational complexity, a customized greedy search strategy is designed as an efficient alternative. 
  Second, to mitigate the temporal discontinuity caused by sparse keyframe sampling, we further introduce interleaved textual \textit{narratives} generated from non-keyframes using off-the-shelf captioners. These narratives are inserted between keyframes based on their true temporal order, forming a coherent and compact representation. 
  Nar-KFC thus serves as a temporal- and content-aware compression strategy that complements visual and textual modalities. 
  Experimental results on multiple long-video benchmarks demonstrate that Nar-KFC significantly improves the performance of popular MLLMs.
  Code will be made publicly available.
\end{abstract}
\vspace{-3mm}

%----------------------------------------------------------------------------
\section{Introduction}
\label{sec:intro}

Large Language Models (LLMs)~\cite{team2024gemma,touvron2023llama2} have demonstrated remarkable progress in human language processing~\cite{bommasani2021opportunities,vaswani2017attention}.
Building upon LLMs, recent advancements in Multimodal Large Language Models (MLLMs)~\cite{liu2023llava,li2024llava_ov,wang2024qwen2vl,chen2024internvl,tong2024cambrian,lin2024vila} have significantly improved open-world visual understanding, such as image captioning~\cite{vinyals2015showandtell} and visual question answering (VQA)~\cite{antol2015vqa}. 
Moving beyond static images, a natural extension of MLLMs is their application to video understanding, and existing studies have also validated their effectiveness in comprehending \emph{short} videos ($\sim$10 s)~\cite{yang2022FrozenBiLM,kim2024IGVLM,yao2024dense}.
However, when scaling MLLMs to %comprehend particularly 
long videos~\cite{fu2024videomme,wu2024longvideobench,chandrasegaran2024hourvideo,zhou2024mlvu} (e.g., hours), several critical challenges emerge.

The primary challenge stems from the inherent context limitation of MLLMs, which cannot accommodate % in effectively perceiving 
the vast volume of visual tokens generated from the whole video.
% which significantly hinders their ability to comprehensive long video understanding.
A prominent solution %to this long-context constraint 
is to extend the context window of language models and fine-tune them on carefully collected long videos.
% fine-tuning
Current video-oriented LLMs, known as VideoLLMs~\cite{lin2024video_llava,jin2024chat-univi,song2024moviechat,xu2024pllava,shu2024videoxl,wang2024retake,chen2024longvila,shen2024longvu,zohar2024apollo}, typically undergo post-training on existing LLMs/MLLMs through: 1) employing a uniform sampling scheme with a relatively large stride, and 2) incorporating token-level merging or compression techniques to enable broader temporal coverage.
However, uniform sampling often fails to preserve key moments relevant to specific instructions, while feeding an excessive number of frames as input introduces redundancy, leading to substantial computational overhead and slower reasoning.
% Training-free
An alternative solution follows a training-free paradigm~\cite{zhang2024LLoVi,kahatapitiya2024languagerepo,wang2024videoagent,wang2024videotree,ma2024drvideo,park2024LVNet}, where raw videos are first converted into sequential captions, which are subsequently processed using the long-range reasoning abilities of LLMs~\cite{achiam2023gpt4}. 
Compared to direct video frame encoding, textual captions inherently require far fewer tokens, allowing efficient inference in a single forward pass. 
Nonetheless, the translation from video frame to caption inevitably results in critical information loss (e.g., important visual features), potentially leading to hallucinated answers caused by the bias of the LLM.
%aimless or biased surmise when answering  questions.  

% 
Building upon the aforementioned paradigms, a fundamental question arises: \textit{Are current MLLMs fully equipped to comprehend long videos despite their limited context length?}
Recent studies have focused on identifying the most query-relevant keyframes~\cite{yu2023selfchain,yuframe-voyager,hu2025mllm-selector,yao2025generative} to enhance inference with MLLMs (with limited context lengths). 
However, due to the high temporal redundancy among adjacent frames, conventional similarity-based keyframe selection often retrieve frames located within narrow time windows, thereby compromising accuracy. 
To address this, techniques such as adaptive keyframe sampling~\cite{tang2025AKS}, inverse transform sampling~\cite{liu2025bolt}, DPP sampling~\cite{sun2025mdp3} have been introduced to promote content diversity during keyframe selection.
Despite a decent boost over existing MLLMs, 
% \textcolor{red}{these methods do not integrate and optimize these two objectives jointly}, 
these methods typically rely on handcrafted or heuristic strategies with limited principled formulations, 
and  empirically, the retrieved frames can be temporally distant, especially in long videos.
As a result, the selection of keyframes introduces temporal discontinuities in information presented to the MLLM, which hinders its understanding of the whole video.

% our work
In this paper, we propose \textit{\textbf{Nar-KFC}}, \textbf{Nar}rating \textbf{K}ey\textbf{F}rames \textbf{C}aptured for training-free long video understanding with MLLMs.
Unlike previous approaches, Nar-KFC jointly considers \textit{query-relevance}, \textit{frame-diversity} and \textit{temporal-continuity} through two main collaborative stages.
% KFC
The first stage \textbf{KFC} 
selects key frames by considering both query relevance and frame diversity, so as to resolve the 
issues of critical information loss from uniform sampling and the too-narrow focus using just query-relevance. We consider key frame selection as a graph problem, where each node is a frame and the edge weight (score) between nodes combines query-relevant similarities and frame-to-frame dissimilarities (frame-diversity). The optimal keyframes are obtained by finding the subgraph with largest total edge weight, which can be formulated
%resolves the critical information loss caused by uniform sampling and the narrow locating problem resulting from query-relevance only selection.
%Specifically, KFC uses a standard VLM model (e.g., CLIP~\cite{radford2021CLIP} or SigLIP~\cite{zhai2023SigLIP}) to extract query and frame embeddings, which are then used to optimize a unified score matrix by combining query-relevant similarities and frame-to-frame dissimilarities (frame-diversity). 
%This optimization is thus formulated as 
as an integer quadratic programming (IQP) problem.
%aimed at selecting keyframes with the highest scores in the unified matrix.
% While advanced optimization techniques can constrain the search space, performance tends to be unstable as problem size increases. 
However, since IQP is NP-hard with exponential complexity, finding exact solutions is infeasible in practice.
To overcome this, we introduce a robust and efficient greedy search (GS) strategy, 
%that iteratively selects frames contributing most to the current frame set.
which, with proper preprocessing of the score matrix, %the greedy algorithm a
achieves 
near-optimal performance with significantly reduced computational complexity.
The second stage \textbf{Nar-KFC} addresses the problem of temporal discontinuities caused when selecting key frames at uneven timesteps. 
% Nar-KFC
%Keyframes optimized via KFC provide broad temporal coverage and ensure query-relevant content for MLLM inference.
%Nonetheless, selecting frames at uneven timestamps disrupts the sequential information flow of video, as information is missing between key frames. This problem especially manifests in long videos where frames might be temporally distant. Such temporal discontinuity can hinder MLLMs from accurately modeling frame-to-frame relationships.
%
%Inspired by the efficiency and global perception advantages of previous caption-only video understanding methods, we propose \textbf{Nar-KFC} in the second stage.
%Generally, 
Specifically, Nar-KFC works by threading keyframes (visual tokens) with \emph{non-keyframe narratives} (text tokens), generated by captioning the intermediate, unselected frames in between, aiming to reconstruct the video as a continuous and coherent sequence in both textual and visual modalities. 
A narrative interval is further applied to control the total number of captions and to reduce the similarity between neighboring descriptions.
Leveraging only a lightweight 2B captioning model, e.g., Qwen2-VL-2B~\cite{wang2024qwen2vl}, Nar-KFC demonstrates significant improvements over existing MLLMs.
% on multiple long-video benchmarks.

% Contributions
In summary, the contributions of this paper are three-fold:
\begin{compactitem}
    \item Jointly considering query-relevance and frame-diversity, we formulate the keyframe capturing process (KFC) in long videos as a subgraph selection problem, implemented as an integer quadratic programming problem.
    %with specific mathematical formulations.
    We introduce a customized greedy search algorithm to solve this problem with significantly reduced and practical time complexity.
    \item We propose Nar-KFC, which threads the optimized keyframes with non-keyframe narratives. 
    By interleaving the two modalities in a temporally continuous manner, Nar-KFC constructs coherent and compact video representations, enabling a broader video coverage under the constraint of frame length limitations in current MLLMs.
    \item Our KFC and Nar-KFC are generally compatible with many MLLMs, achieving consistent improvements across four mainstream MLLMs on multiple long-video benchmarks.
    % in a training-free mechanism. 
\end{compactitem}
\vspace{-2mm}

%----------------------------------------------------------------------------

\section{Related work}
\label{sec:related}
% LLM -> MLLM -> MLLM for video
Transformer-based LLMs have revolutionized the field of natural language processing~\cite{brown2020language,openai2023chatgpt,grattafiori2024llama3,achiam2023gpt4}.
By incorporating multimodal inputs such as images and videos~\cite{li2024llava_ov,zhu2023minigpt4} with a vision encoder, e.g., ViT~\cite{dosovitskiy2020ViT}, researchers further extend powerful LLMs to multimodal large language models (MLLMs) for open-world visual understanding~\cite{alayrac2022flamingo,li2023blip2,liu2023llava}.
Despite similar advancements of MLLMs on various video understanding tasks including video captioning~\cite{chen2024sharegpt4video,yang2023vid2seq,wu2024dibs}, video question answering~\cite{maaz2023videochatgpt,li2023videochat,min2024morevqa}, and temporal reasoning~\cite{qian2024momentor}, significant challenges emerge when scaling to long videos due to the substantial amount of video frames not fitting in the limited context length of LLMs~\cite{wu2024longvideobench}.

Recent studies have explored methods to extend the context length of LLMs~\cite{wan2024efficient,xiong2024effective}, or introduced various token-level merging and compression techniques~\cite{song2024moviechat,shen2024longvu,li2024llama-vid,shu2024videoxl,wang2024retake} to accommodate more frames as input. However, these approaches typically require additional fine-tuning of existing language models, which increases computational complexity and introduces the risk of hallucinations~\cite{liu2024lost}. 
Given that textual tokens are significantly fewer than visual frames, another line of research first converts all video frames into textual descriptions, which are then used for long video inference, either by summarizing them~\cite{zhang2024LLoVi,park2024LVNet} or identifying central frames based on textual similarity via agents~\cite{wang2024videoagent,wang2024videotree,ma2024drvideo}. Nonetheless, the converting process inevitably leads to critical information loss, thereby compromising performance.
Other studies, while maintaining the number of input frames, adopt alternative sampling strategies instead of default uniform sampling to obtain higher-quality frames for input.
In general, query relevance is the primary criterion for selecting frames that are semantically closest to the query~\cite{yu2023selfchain,lin2024vila,wang2024videoagent,wang2024weakly}.
Methods such as AKS~\cite{tang2025AKS}, BOLT~\cite{liu2025bolt}, Frame-Voyager~\cite{yuframe-voyager} further propose adaptive sampling, inverse transform sampling, and optimal frame combination sampling to identify keyframes that are both query-relevant and temporally distinctive. 
Nevertheless, the methods often rely on manually designed heuristics without principled theoretical guidance,
and the selected keyframes are often undistributed and distant over long intervals, especially in hours-long videos (e.g., 3600 frames per hour at 1 fps). This temporal sparsity weakens the relationships between frames and can cause confusion in MLLM inference. 

In contrast to previous works, we formulate long video keyframe selection as a graph-based optimization problem with a clearly defined objective, and further leverage the efficiency of textual descriptions.
% we leverage the strengths of both keyframes selection in long videos and the efficiency of textual descriptions. 
Our approach jointly considers query relevance, content diversity, and temporal continuity, aiming to construct optimal combinations of keyframes with interleaved narratives, under the constraints of MLLM context length.
% \textcolor{red}{...}
\vspace{-1.5mm}

%----------------------------------------------------------------------------
\section{Method}
\label{sec:method}
\vspace{-3mm}
In this section, we present the two stages of our training-free long video understanding approach using MLLMs: KFC and Nar-KFC. 
\vspace{-1.5mm}
% Roadmap
% In this section, we present \textit{Threading Keyframe with Narratives} method to enhance MLLM-oriented long video understanding. 
% It consists of two main components: 1) \textbf{capturing keyframes} (\textbf{KFC}), discussed in Section~\ref{subsec:kfc}, and 2) \textbf{narrating the keyframes captured} (\textbf{Nar-KFC}), in Section~\ref{subsec:nar-kfc}.

\subsection{KFC: KeyFrame Capturing}
\label{subsec:kfc}

\begin{wrapfigure}{r}{0.4\textwidth}
\vspace{-14mm}
  \centering
  \includegraphics[width=0.4\textwidth]{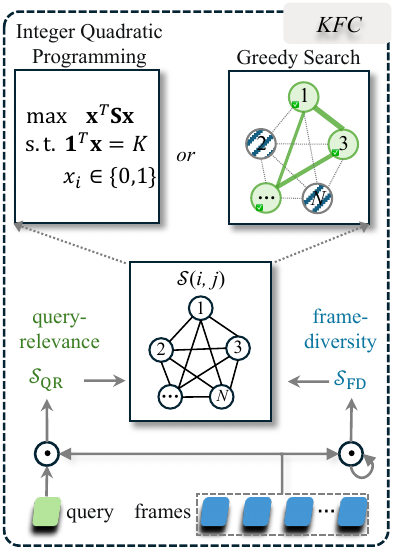}
  \caption{Illustration of keyframe capturing (KFC). $\mathcal{S}_{\mathrm{QR}}$ and $\mathcal{S}_{\mathrm{FD}}$ scores are computed via inner dot production.}
  \label{fig:kfc}
  \vspace{-6mm}
\end{wrapfigure}

% Trivial uniform sampling is commonly used in previous \textit{short} video understanding, as it provides a reasonably consistent temporal structure for video representations. However, in the context of long videos, uniform sampling can overlook significant information with limited input.
Uniform sampling is commonly used in \textit{short} video understanding for consistent temporal structure. However, for \textit{long} videos, it often misses important information with limited input.
%
% Although recent works~\cite{} have demonstrated the importance of selecting query-relevant frames for long video QA, they tend to overlook the narrow locating problem from simple query-relevance selection due to the high-similar character among adjacent video frames, as discussed in 
% Section~\ref{sec:intro}.
While recent works emphasize selecting query-relevant frames for long video QA, they tend to overlook the problem of narrow focus 
%narrow locating problem 
due to the high similarity between adjacent frames.
%, as discussed in Section~\ref{sec:intro}.
%
To address this, we first propose a keyframe capturing method that simultaneously considers query-relevance and frame-diversity, modeling the selection process as subgraph selection problem, which is implemented as a theoretical optimization problem.

\textbf{Preliminaries.} 
General video understanding tasks, e.g., video summarization and grounding~\cite{liu2024timecraft, xiao2024can}and long-video QA, can be similarly formulated as $(V, q) \rightarrow \mathrm{Answer}$, where $V=\{f_i\}_{i=1}^N$ represents a video with $N$ frames, $f_i$ is the $i$-th frame, and $q$ is the query. 
Considering an MLLM model as a neural function $\mathcal{M}(\cdot)$ with its limited contextual perceiving length, the normal video QA process reasoned by an MLLM model can be formulated as $\mathcal{M}(\{f_i\}_{i=1}^K, q)\rightarrow \mathrm{Answer}, 1\leq K\ll N$, meaning that only $K$ frames are captured for representing video $V$. We next consider two criteria for selecting the $K$ frames, query-relevance and frame-diversity.

\textbf{Query-relevance.} 
Since different questions can be asked on the single video, it is crucial to identify frames that correspond to a specific query first, especially in long videos.
Here, a standard two-stream vision-language model (VLM), e.g., CLIP~\cite{radford2021CLIP}, is used to extract  embeddings $\{\mathbf{f}_i\}_{i=1}^N$ and $\mathbf{q}$ for the frames and the query, respectively. 
After standard normalization of all embeddings, the query-relevance score $\mathcal{S}_{\mathrm{QR}}$ is computed as the cosine similarity between the two, $\mathcal{S}_{\mathrm{QR}}(i)=\mathrm{sim}(\mathbf{f}_i, \mathbf{q})$. 
%$ \in \mathbb{R}^{N\times1}$.

\textbf{Frame-diversity.}
To avoid retrieving query-relevant only frames that are narrowly located in a small time range, we explicitly encourage diversified content when choosing the $K$ frames.
%, expecting the frames selected should be different from each other within the candidate set.
In particular, we use the \textit{inverse} of cosine similarity between every pair of frame embeddings (normalized) to represent the diversity score. The function $\mathrm{exp(\cdot)}$ is applied to constrain the score between 0 and 1, formulated as $\mathcal{S}_{\mathrm{FD}}(i,j)=\mathrm{exp}(-\mathrm{sim}(\mathbf{f}_i, \mathbf{f}_j))$. 
%, 1\leq i,j\leq N$, where $\mathcal{S}_{\mathrm{FD}}\in\mathbb{R}^{N\times N}$.

\textbf{Objective.} 
The final score combines $\mathcal{S}_{\mathrm{QR}}$ and $\mathcal{S}_{\mathrm{FD}}$ to jointly identify keyframes that are both query-relevant and temporally diverse for KFC,
\begin{equation}
    \mathcal{S}(i,j) = \mathcal{S}_{\mathrm{QR}}(i) + \alpha \mathcal{S}_{\mathrm{FD}}(i,j) = 
    \mathrm{sim}(\mathbf{f}_i, \mathbf{q}) + \alpha \mathrm{exp}(-\mathrm{sim}(\mathbf{f}_i, \mathbf{f}_j)), % \  1\leq i, j\leq N,
    \label{eq:score}
\end{equation}
 where $\alpha$ is a weighting factor that controls the relative importance of the two terms. 

%To achieve this, we expand $\mathcal{S}_{\mathrm{QR}}$ to a matrix of size $N\times N$ dimension (assuming there are $N$ identical queries), and then aggregate the two scores to obtain the final objective score matrix $\mathcal{S}$:
%\begin{equation}
%    \mathcal{S} = \mathrm{expand}(\mathcal{S}_{\mathrm{QR}}) + \alpha \mathcal{S}_{\mathrm{FD}} = 
 %   \mathrm{expand}(sim(\mathbf{f}_i, \mathbf{q})) + \alpha\cdot \mathrm{exp}(-sim(\mathbf{f}_i, \mathbf{f}_j)), \  1\leq i, j\leq N,
 %   \label{eq:score}
%\end{equation}
%where $\mathcal{S}\in\mathbb{R}^{N\times N}$, and $\alpha$ is a weighting factor that controls the relative importance of $\mathcal{S}_{\mathrm{FD}}$ and balances the two components. 
%
Next, 
%For simplicity, we consider only the upper triangle of $\mathcal{S}$, 
as illustrated in Fig.\ref{fig:kfc}, 
we construct a graph where each node is a frame, and the edge weight between node pair $(i,j)$ is $\mathcal{S}(i,j)$.
%where each entry $\mathcal{S}_{ij}$ represents the relevance of $i$-th frame to the query and the dissimilarity between $i$-th and $j$-th frames. 
The selection of $K$ keyframes can then be cast as a subgraph selection problem with 
%This reformulates 
the original objective as follows: ``\textit{given $N$ nodes (frames), construct a sub-graph by selecting $K$ nodes (keyframes) so as to maximize the total edge weight of the subgraph.}'' Mathematically, this objective can be expressed as the optimization problem:
    \begin{align}
    \max_{Y \subset \{1,\cdots,N\}, |Y|=K} \sum_{(i,j) \in \mathcal{I}} \mathcal{S}(i,j), 
    \label{eqn:subgraph}
    \end{align}
where $Y=\{y_1,\cdots,y_K\}$ is the index set of the $K$ keyframes and $\mathcal{I}$ denotes all edges of pairs $(i, j)$. 

%"\textit{Given a video with $N$ frames, select $K$ keyframes that maximize the summed scores in $\mathcal{S}$}."

\begin{figure}
  \centering
  % \fbox{\rule[-.5cm]{0cm}{4cm} \rule[-.5cm]{4cm}{0cm}}
  \includegraphics[width=0.95\textwidth]{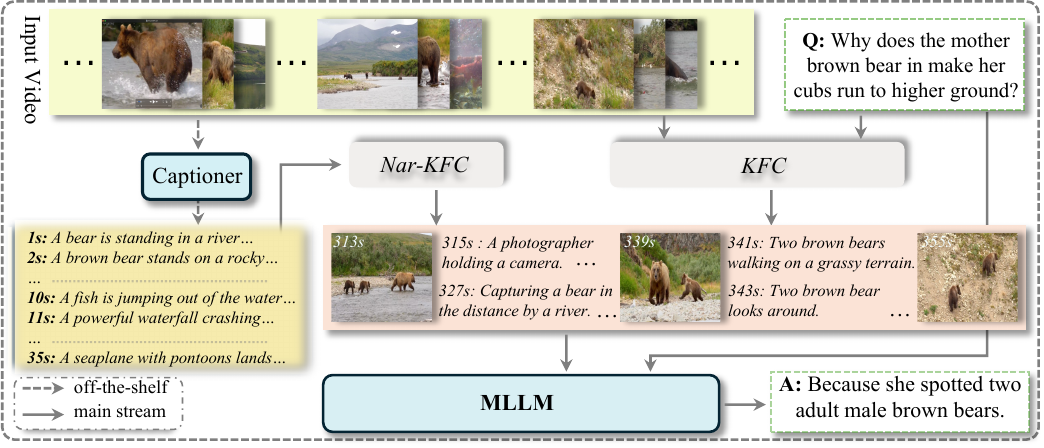}
  \vspace{-2mm}
  \caption{Illustration of Nar-KFC. We represent long videos by threading KFC-optimized keyframes with temporally interleaved narratives, where the narratives are generated frame-wise by an off-the-shelf captioner. Nar-KFC constructs a continuous representation to facilitate MLLM inference.}
  \label{fig:nar-kfc}
  \vspace{-3mm}
\end{figure}

\subsubsection{Optimal solution: Integer Quadratic Programming}
\label{sec:QP}
Our objective closely resembles the classic Knapsack problem~\cite{salkin1975knapsack}, which can be commonly solved by dynamic programming or integer linear programming.
The problem in (\ref{eqn:subgraph}) can be rewritten equivalently as 
%For ease of understanding, we model the above objective as 
an \textbf{integer quadratic programming (IQP)} problem, % defined as:
\begin{equation}
%    \begin{aligned}
    \max_{\mathbf{x}} \ \  \mathbf{x}^T \mathbf{S}\mathbf{x} \quad
    \mathrm{s.t.}\ \ \mathbf{1}^T\mathbf{x}=K,\ x_i\in\{0,1\},
%    \end{aligned}
\label{eq:iqp}
\end{equation}
where $x_i=1$ indicates that the $i$-th frame is selected, $\mathbf{x}=[x_1, x_2, \cdots, x_N]^T$, and $\mathbf{S} \in \mathbb{R}^{N \times N}$ is the score matrix with $\mathbf{S}_{i,j} = \mathcal{S}(i,j)$ for $i <  j$, and $\mathbf{S}_{i,j}=0$ otherwise.
Here, only the upper triangle of $\mathbf{S}$ is considered. A discussion of symmetrical $\mathbf{S}$ is detailed in Appendix \S\ref{sec:sup_symmtric}.
The search space is $C(N, K)=\frac{N!}{K!(N-K)!}$, and the time complexity of solving IQP is exponential regardless of whether the objective is convex or non-convex, making it impractical to get exact solutions in real cases. 
Modern optimization tools, e.g., CPLEX~\cite{bliek1u2014cplex}, typically address this by 
relaxing the binary constraint and allowing $x_i\in[0,1]$, converting the problem into a continuous optimization task.
Solutions can then be obtained using methods like interior-point or Lagrange multiplier methods, with a complexity of $O(N^3)$.
Subsequently, the Branch \& Bound algorithm~\cite{morrison2016branch} is applied to prune the search space and retrieve optimal integer solutions of $x_i$, but the worst-case time complexity remains exponential.
%. However, the worst-case time complexity of this step remains exponential.

\subsubsection{Near-optimal solution: Greedy Search}
\label{sec:GS}
Solving the IQP optimally is computationally prohibitive for large $N$, e.g., long videos with thousands of frames.
 Thus,
%To mitigate the exponential complexity of solving the IQP optimally in practice, 
we propose a customized \textbf{greedy search (GS)} strategy that provides a robust and near-optimal approximation to the original objective.
We first pre-process the score matrix to reduce noise across adjacent columns/rows, and shrinks the problem size for improved computational efficiency.
Specifically, we apply singular value decomposition (SVD) to the score matrix $\mathbf{S}$, retaining the top $r$ singular values to construct a low-rank approximation $\mathbf{S}_r\in \mathbb{R}^{N\times N}$.
This matrix is then uniformly downsampled to $\mathbf{S}_{rd}\in\mathbb{R}^{\frac{N}{d}\times\frac{N}{d}}$ with a downsampling ratio $d$. 
The GS algorithm begins by selecting the most query-relevant frame as the starting point.
It then iteratively adds the frame with the highest \textit{cumulative} score when compared to the already selected frames.
In the final refinement step, the algorithm examines the $k$-nearest neighbors of each selected frame $y_i$, replacing $y_i$ with a neighboring frame if it yields a higher cumulative score based on $\mathcal{S}_r$.  
A summary of the algorithm is provided in Alg.~\ref{algo:GS}, 
and its overall time complexity is $O(NK)$. %, and the entire process is outlined i
\begin{algorithm}
\caption{The Pseudocode of Near-optimal solution: Greedy Search}
\label{algo:GS}
\KwIn{Query-relevant score $\mathcal{S}_{\mathrm{QR}}$,  score matrix $\mathbf{S}$, number of retained singular values $r$, downsample ratio $d$, number of frames $N$, number of keyframes $K$, neighbor window $k$.}
\KwOut{Indices of selected $K$ frames set $Y=\{y_1, y_2, \cdots, y_K\}$}
$\mathbf{S}_r\leftarrow\mathrm{LowRank}(\mathbf{S})$; \quad $\mathbf{S}_{rd}\leftarrow\mathrm{Downsample}(\mathbf{S}_r, d)$; \quad \tcp{\small Decompose and downsample $\mathbf{S}$}
$y_1 = \mathop{\mathrm{argmax}}_i  \mathcal{S}_{\mathrm{QR}}(i)$. \ $Y \leftarrow \{y_1\}$ \ \  \tcp{\small Initialize with the most query-relevant frame}
\For{$i \gets 2$ \KwTo $K$}{   
    \For{$j \gets 1$ \KwTo $N$}{
    $y_i = \mathop{\mathrm{argmax}}_j \ 
    \sum_{y\in Y} S_{rd}(y,y_j)$ \ \tcp{\small Select frame with highest sum}
    %sum(Y, y_j|\mathbf{S}_{rd})$; 
    $Y\leftarrow Y \cup y_i$
    }
}
\For{$i \gets 1$ \KwTo $K$}{
    $y_i = \mathrm{Refine}(y_i, k|\mathbf{S}_r)$; \quad \tcp{\small Refine selection within $k$-nearest neighbors}
}
\Return{$Y = \mathrm{sorted}\{y_1, y_2, \cdots, y_K\}$}\;
\end{algorithm}
\vspace{-5mm}

\subsection{Nar-KFC: Threading \textit{Keyframe} with \textit{Narratives}}
\label{subsec:nar-kfc}
Keyframes captured by KFC significantly enhance the performance of  MLLMs compared to the default uniform inference mechanism.
% effectively addressing both the critical information loss and the narrow locating problem.
However, it overlooks the issue of \textit{temporal-continuity} in frame sequences. 
% Due to the severely uneven distribution of selected frames, the temporal relationships among them tend to be weak, which can lead to confusion during model inference.
Due to the severely uneven distribution of selected frames, temporal relationships become weak, often leading to confusion during inference.
Currently, no mature solution exists at either the token or frame level, apart from the brute-force approach of densely adding more frames.

To this end, we propose \textbf{Nar-KFC}, which threads keyframes with text narratives to construct a continuous and coherent input in an interleaved form.
Specifically, we first use a lightweight off-the-shelf captioner, e.g., Qwen2-VL-2B~\cite{wang2024qwen2vl}, to generate captions $\{c_i\}_{i=1}^N$ for each frame using a simple prompt as "\texttt{<USER> Describe this video frame in no more than 15 words.}"
Given the unevenly distributed keyframes $\{f_{y_i}\}_{i=1}^K$ from KFC, we insert \textit{captions from non-keyframes} between the keyframes, arranging them according to their true temporal order. 
Each $y_i$ denotes the timestamp, and a uniform interval $\bigtriangleup$ is set between captions to control the total number of inserted narratives. 
The overall long video inference to a MLLM model $\mathcal{M}$ is formulated as:
\begin{equation}
    \mathcal{M}\big(\{f_{y_1}, c_{y_1+\bigtriangleup}, \cdots, c_{y_2-\bigtriangleup}, f_{y_2}, c_{y_2+\bigtriangleup}, \cdots, c_{y_K-\bigtriangleup}, f_{y_K}\}, q\big)\rightarrow \mathrm{Answer}.
\end{equation}

\textbf{Viability} of Nar-KFC.
MLLMs are typically trained via instruction tuning on both visual and textual modalities, making them well-suited to process our interleaved inputs of keyframes and narratives.
\\
\textbf{Rationality} of Nar-KFC.
The approach provides supplemental information for unsampled frames, enabling a more temporally continuous input that helps MLLMs ``narrate'' the story between keyframes.
From another perspective, Nar-KFC can be seen as a form of compression — retaining only the most informative keyframes, which are encoded into fine-grained visual tokens, while representing less critical segments with brief textual descriptions. 
This complementary two-stream mechanism is analogous to methods like Two-Stream~\cite{simonyan2014two-stream} (combining RGB frames with optical flow), 
or SlowFast~\cite{feichtenhofer2019slowfast} and
SlowFast-LLaVA~\cite{xu2024slowfast-llava}, where the caption stream functions as \textit{a fast branch} that traverses a wider temporal range (as in the low frame rate of the slow branch in SlowFast).
These mechanisms together help explain the effectiveness of Nar-KFC in (long) video understanding.

%----------------------------------------------------------------------------
\section{Experiments}
\label{sec:experiment}
\vspace{-2.5mm}
We next present experiments to demonstrate the efficacy of our Nar-KFC.

%\subsection{Experiment Settings}
%\label{subsec:ex_setup}
\textbf{Evaluation Benchmarks.}
We evaluate on three recent  long-video benchmarks: 1) \textit{Video-MME}~\cite{fu2024videomme}, consisting of 2,700 human-annotated QA pairs, with an average video duration of 17 min;
2) \textit{LongVideoBench}~\cite{wu2024longvideobench}, using the validation set (denoted as \emph{LVB}), which contains 1,337 QA pairs with average duration of 12 min;
3) \textit{MLVU}~\cite{zhou2024mlvu}, a multi-task benchmark, where we use the multiple-choice task (M-avg), comprising 2,593 questions across 9 categories, with average  duration of 12 min. \\
\textbf{Evaluation Models.} 
We consider three advanced MLLMs, including InternVL2~\cite{chen2024internvl}, Qwen2-VL~\cite{wang2024qwen2vl}, LLaVA-OneVision~\cite{li2024llava_ov}, as well as one state-of-the-art VideoLLM, i.e., LLaVA-Video~\cite{zhang2024llava_video}, to verify the effectiveness of our method. 
All models use their 7B (or 8B) parameter variants.
We re-implement baseline results (uniform sampling) of these MLLMs using VLMEvalKit~\cite{duan2024vlmevalkit}, which may yield slight differences compared to other public toolkits, e.g., LMMs-Eval~\cite{lmms_eval2024}.\\
\textbf{Implementation Details.}
We use CLIP-ViT-L-336px~\cite{radford2021CLIP} %in default for 
to extract query and video frame embeddings. To reduce computational costs, candidate frames are sampled from raw videos at 1 frame per second.
The balancing coefficient $\alpha$ is set to 1 in (\ref{eq:score}).
For solving the quadratic integer programming, we limit the maximum search nodes to 40k in CPLEX~\cite{bliek1u2014cplex}.
In our customized greedy search algorithm, we retain the top $\frac{N}{4}$ singular values to form the low-rank approximation of  the  score matrix $\mathbf{S}$ and further downsample it to a fixed resolution of 128$\times$128.
%, where $N$ denotes the size of $\mathcal{S}$.
The refinement window size $k$ is set to 2 (see ablations in Appendix \S\ref{sec:sup_neighbor_k}), i.e., 4 neighbor frames are examined during refinement.
Unless otherwise stated, all ablations are conducted using the InternVL2 model on the Video-MME benchmark.
Experiments are run on 8 A100 (40GB VRAM) GPUs.

\vspace{-4mm}
\subsection{Benchmark Results}
\vspace{-2mm}

\textbf{Comparisons with State-of-the-Arts.}
We conduct comprehensive comparisons between our approach and several recent MLLMs and VideoLLMs with default $K=8$ frames, as shown in Tab.~\ref{tab:sota1}.
Earlier works, e.g., Video-LLaVA~\cite{lin2024video_llava}, VideoChat2~\cite{li2023videochat}, Chat-UniVi-V1.5~\cite{jin2024chat-univi}, VideoLLaMA2~\cite{cheng2024videollama},
% ShareGPT4Video~\cite{chen2024sharegpt4video}, 
\etc, are omitted here and fully reported in Appendix \S\ref{sec:sup_full_sota}.
% are omitted due to space constraints and their relatively lower performance (see Appendix \S\ref{sec:sup_full_sota} for full results).
%
Our methods, KFC and Nar-KFC, deliver consistent and significant gain over four baselines across three long-video benchmarks.
% Video-MME
%Specifically, on the 
On \textit{Video-MME} (no sub.), Nar-KFC outperforms four MLLM baselines by 4.4\%, 1.7\%, 4.5\%, and 5.7\%, respectively.
Using the strongest baseline, i.e., LLaVA-Video, Nar-KFC achieves state-of-the-art performance (61.6\%), surpassing previous VideoLLMs - even those using larger LLMs (e.g., VILA-34B, 58.3\%) or more frames (e.g., Video-XL$^{\mathrm{256frm}}$, 55.5\%). 
Incorporating larger numbers of frames may introduce noise and irrelevant information, which can be well addressed by our keyframe capturing and narrating strategies.
% Longvideobench
%As for the
On \textit{LVB}, our method also achieves notable performance improvements, e.g., 52.3\% \textit{vs.} 53.9\% with InternVL2 and  53.4\% \textit{vs.} 54.6\% with Qwen2-VL, although the overall gain is partly offset by videos shorter than 1 min. Nevertheless, the improvement on typical long videos ($\sim$1 hour, as shown in parentheses) reaches 3.5\% and 2.9\%, respectively, demonstrating clear advantages in long video understanding.
% MLVU
On \textit{MLVU}, our KFC-only strategy (without narrations) yields an average improvement of over 6\% across four MLLMs. The use of query-relevant and diverse keyframes significantly boosts performance on Needle-in-a-haystack~\cite{zhang2024needle} and counting questions in MLVU. Furthermore, appending narratives provides additional and robust gains by preserving temporal continuity.
Detailed analysis are further presented in Appendix \S\ref{sec:sup_mlvu_analysis}.
We provide more results of our methods on relatively short EgoSchema~\cite{mangalam2023egoschema} and NExTQA~\cite{xiao2021nextqa} benchmarks in Appendix \S\ref{sec:sup_sota}.

\begin{table}
  \caption{Comparisons with previous VideoLLMs on three common long-video benchmarks: Video-MME, LVB, and MLVU.
  For Video-MME, we report performance with two standard settings: without subtitles (no sub.) and with subtitles (sub.).
  LVB denotes the LongVideoBench validation set, with results for (15m, 60m] long videos shown in parentheses.
  Methods that use significantly more frames and larger-sized LLM are marked in \textcolor{gray}{gray}.
  The reported results are accuracy percentage.
  %Accuracy sign \% is omitted for clarity.
  }
  \label{tab:sota1}
  \centering
  \setlength{\tabcolsep}{5pt}  % 缩小列间距
  \resizebox{\textwidth}{!}{
  \begin{tabular}{lccccccc}
    \toprule
    \multirow{2}{*}{Model} & \multirow{2}{*}{Size} & \multicolumn{4} {c}{Video-MME$_{\mathrm{(no\ sub.\ /\ sub.)}}$} & \multirow{2}{*}{LVB} & \multirow{2}{*}{MLVU}\\
    \cmidrule(r){3-6}
    & & Short    & Medium     & Long & Overall$_{\sim 17m}$ & $_{\sim 12m}$  & $_{\sim 12m}$ \\
    \hline
    VILA~\cite{lin2024vila} & 8B & 57.8 / 61.6 & 44.3 / 46.2 & 40.3 / 42.1 & 47.5 / 50.0 & - & 46.3 \\
    LLaVA-NeXT-QW2~\cite{liu2024llavanext} & 7B & 58.0 /\pos  & 47.0 /\pos & 43.4 /\pos & 49.5 /\pos & {-} & {-} \\
    MiniCPM-V2.6~\cite{yao2024minicpm} & 7B  & 61.1 / 63.8 & 50.3 / 50.2 & 46.4 / 45.4 & 52.6 / 53.1 & 51.2 & 55.4 \\
    Frame-Voyager~\cite{yuframe-voyager} & 8B & 67.3 /\pos & 56.3 /\pos & 48.9 /\pos & 57.5 /\pos & - & 65.6 \\
    % LongVU~\cite{shen2024longvu} & 7B & 64.7 /\pos & 58.2 /\pos & 59.5 /\pos & 60.6 /\pos & - & 65.4 \\
    \cg LongVILA$^{\mathrm{256frm}}$~\cite{chen2024longvila} & \cg 8B & \cg 61.8 /\pos & \cg 49.7 /\pos & \cg 39.7 /\pos & \cg 50.5 /\pos & \cg - & \cg -\\
    \cg Video-XL$^{\mathrm{256frm}}$~\cite{shu2024videoxl} & \cg 7B & \cg 64.0 / 67.4 & \cg 53.2 / 60.7 &\cg 49.2 / 54.9 &\cg 55.5 / 61.0 &\cg 50.7 &\cg 64.9 \\
    \cg LLaVA-NeXT-Video~\cite{zhang2024llavanextvideo} &\cg 34B &\cg  61.7 / 65.1 &\cg 50.1 / 52.2 &\cg 44.3 / 47.2 &\cg 52.0 / 54.9 &\cg 50.5 &\cg 58.8 \\
    \cg VILA~\cite{lin2024vila} &\cg 34B &\cg 70.3 / 73.1 &\cg 58.3 / 62.7 &\cg 51.2 / 55.7 &\cg 58.3 / 61.6 &\cg - &\cg 57.8 \\
    \hline
    InternVL2~\cite{chen2024internvl} & 8B & 62.1 / 63.9 & 48.2 / 48.7 & 45.2 / 44.9 & 51.9 / 52.5 & 52.3 (45.2) & 54.3 \\
    \rowcolor{lighterblue}
    \quad \textbf{+ KFC} & 8B & 64.3 / 65.4 & 49.6 / 52.3 & 46.1 / 47.3 & 53.1 / 55.0 & 53.3 (47.2) & 62.2 \\
    \rowcolor{lightblue}
    \quad \textbf{+ Nar-KFC} & 8B & 67.2 / 67.7 & 54.7 / 57.9 & 47.1 / 48.9 & \textbf{56.3} / \textbf{58.1} & \textbf{53.9} (\textbf{48.8}) & \textbf{64.4} \\
    \hline
    Qwen2-VL~\cite{wang2024qwen2vl} & 7B & 65.7 / 66.9 & 52.8 / 53.0 & 46.7 / 48.6 & 55.0 / 56.1 & 53.4 (45.0) & 59.6\\
    \rowcolor{lighterblue}
    \quad \textbf{+ KFC} & 7B & 68.2 / 69.7 & 53.3 / 54.9 & 48.4 / 50.2 & \textbf{56.7} / \textbf{58.3} & \textbf{54.6} (\textbf{47.9}) & 65.9 \\
    \rowcolor{lightblue}
    \quad \textbf{+ Nar-KFC} & 7B & 68.8 / 69.3 & 53.4 / 55.3 & 48.0 / 49.0 & \textbf{56.7} / 57.9 & 53.6 (46.3) & \textbf{68.5} \\
    \hline
    LLaVA-OneVision~\cite{li2024llava_ov} & 7B  & 65.2 / 67.1 & 51.7 / 54.4 & 45.1 / 46.1 & 53.3 / 55.9 & 54.5 (45.7) & 58.5 \\
    \rowcolor{lighterblue}
    \quad \textbf{+ KFC} & 7B & 66.4 / 69.1 & 52.9 / 56.8 & 46.8 / 48.8 & 55.4 / 58.2 & 55.6 (47.3) & 65.0 \\
    \rowcolor{lightblue}
    \quad \textbf{+ Nar-KFC} & 7B & 67.2 / 68.6 & 57.1 / 59.8 & 49.1 / 51.0 & \textbf{57.8} / \textbf{59.8} & \textbf{56.5} (\textbf{48.2}) & \textbf{66.2} \\
    \hline
    LLaVA-Video~\cite{zhang2024llava_video} & 7B & 67.2 / 69.4 & 53.2 / 53.4 & 47.2 / 47.3 & 55.9 / 56.7 & 54.2 (46.5) & 60.5 \\
    %  52.7 (43.8)
    \rowcolor{lighterblue}
    \quad \textbf{+ KFC} & 7B & 68.3 / 70.0 & 55.1 / 57.4 & 49.4 / 51.6 & 57.6 / 59.7 & 56.5 (49.3) & 66.9 \\
    \rowcolor{lightblue}
    \quad \textbf{+ Nar-KFC} & 7B & 71.2 / 72.7 & 61.4 / 62.3 & 52.0 / 53.9 & \textbf{61.6} / \textbf{63.0} & \textbf{57.7} (\textbf{50.2}) & \textbf{67.7} \\
    \bottomrule
  \end{tabular}
  }
  \vspace{-5mm}
\end{table}

\begin{figure}
  \centering
  \includegraphics[width=1.0\textwidth]{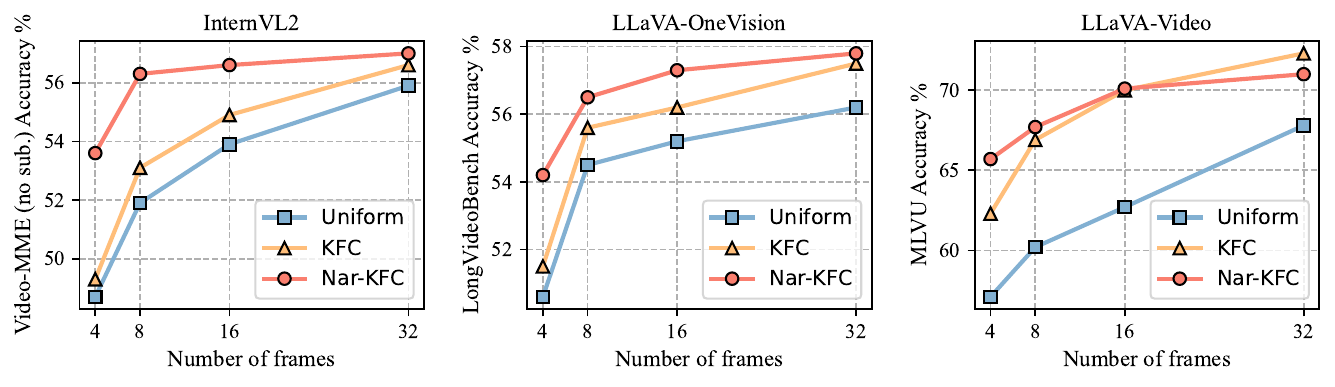}
  \vspace{-7mm}
  \caption{Accuracies (\%) of uniform sampling, KFC, and Nar-KFC versus numbers of keyframes.  }
  \label{fig:frame_num}
  \vspace{-5mm}
\end{figure}

\textbf{Comparisons with varying number of keyframes.}
In Fig.~\ref{fig:frame_num}, we compare the performance of KFC and Nar-KFC against uniform sampling across the three benchmarks  %(Video-MME (no sub.), LVB, and MLVU), 
using InternVL2, LLaVA-OneVision, and LLaVA-Video, respectively.
Due to Qwen2-VL's dynamic resolution mechanism~\cite{wang2024qwen2vl,dehghani2023patch}, increasing the number of keyframes often leads to memory overflow, so its results are omitted.
Across all settings, both KFC and Nar-KFC consistently outperform uniform sampling. 
Notably, Nar-KFC shows substantial gains when the number of keyframes is limited (e.g., 4 or 8), due to its ability to provide broad video coverage via interleaved textual narratives, which enhances holistic video understanding even with sparse visual inputs.
As the number of keyframes increases, the performance gap between uniform sampling and our methods narrows.
This can be attributed to two factors:
%two factors
1) uniform frames have a higher chance of capturing key moments when more frames are used; and 2) many video QA questions typically only require a few number of frames %do not require a typical large number of frames 
to accurately answer in current benchmarks.
Interestingly, on MLVU, KFC alone outperforms Nar-KFC with 32 keyframes, suggesting that when sufficient keyframes are present, the added benefit of narratives diminishes.
These results underscore the strength of KFC in selecting informative keyframes 
% for long videos
while demonstrating that narratives are particularly valuable when MLLMs have limited context capacity.

\begin{table}[t] % ← 告诉 LaTeX 优先放在页面顶部
% \vspace*{-1em}   % ← 减少默认浮动体顶部间距，可调节
\centering
\begin{minipage}{0.53\textwidth}
  \centering
  \captionof{table}{Main component ablation results in Nar-KFC. ``S, M, L'' refer to short, medium, and long video categories in the Video-MME (sub.) benchmark.}
  \vspace{-2mm}
  \label{tab:ablation}
  \vspace{2mm}
  \setlength{\tabcolsep}{3pt}
  \resizebox{\textwidth}{!}{
  \begin{tabular}{lcccccr}
    \toprule
    \multirow{2}{*}{Strategy}  & \multicolumn{4}{c}{Video-MME} &  \multirow{2}{*}{MLVU} &  \multirow{2}{*}{Time} \\
    \cmidrule(r){2-5}
    & S & M & L & Overall & \\
    \midrule
    Uniform & 63.9 & 48.7 & 44.9 & 52.5 & 54.3 & $O(1)$ \\
    \  + \textbf{Nar}ratives & 66.1 & 54.9 & 45.2 & 55.4 & 59.4 & $O(N)$  \\
    \midrule
    \cg \textbf{KFC} (IQP) & \cg 65.9 &\cg 52.9 &\cg 46.4 &\cg 55.1 & \cg 62.0 & \cg $O(2^N)$ \\
    \textbf{KFC} (GS) & 65.4 & 52.3 & 47.3 & 55.0 & 62.2 & $O(NK)$ \\
    \  w/o $\mathcal{S}_{\mathrm{QR}}$ & 62.3 & 47.8 & 45.3 & 51.8 & 57.3 & $O(NK)$ \\
    \  w/o $\mathcal{S}_{\mathrm{FD}}$ & 63.6 & 49.4 & 44.6 & 52.5 & 60.9 & $O(NK)$ \\
    \midrule
    \textbf{Nar-KFC} & \textbf{67.7} & \textbf{57.9} & \textbf{48.9} & \textbf{58.1} & \textbf{64.4} & $O(NK)$ \\
    \bottomrule
  \end{tabular}}
\end{minipage}
\hfill
\begin{minipage}{0.43\textwidth}
  \centering
  \captionof{table}{Effects of including pre-processing and refinement stages in the KFC Greedy Search (GS) method. V-MME denotes the overall Video-MME (sub). 
  % and the overall performance is reported. 
  Line (ii') indicates Downsampling  \textit{without} LowRank. The final KFC (GS) strategy integrates all components from (i) to (iv).}
  \label{tab:gs_module}
  % \vspace{-2mm}
  \setlength{\tabcolsep}{3pt}
  \resizebox{\textwidth}{!}{
  \begin{tabular}{llcc}
    \toprule
    % \multirow{2}{*}{Ex\#} & \multirow{2}{*}{Strategy} & \multicolumn{2}{c}{Video-MME} \\
    % \cmidrule(r){3-4}
    % & & M & Overall \\
    Ex\# & Strategy & V-MME & MLVU \\
    \midrule
     & Vanilla GS & 52.3 & 60.4 \\
    (i) & \ + Initialization & 53.3 & 61.0 \\
    (ii) & \ \ \ + LowRank & 53.7 & 61.8 \\
    (ii') & \ \ \  + Downsample & 53.9 & 61.6 \\
    (iii) & \ \ \ + LowRank + Downsample & 54.7 & \textbf{62.2} \\
    (iv) &  \ \ \ \ \ + Refinement (\textbf{KFC}) & \textbf{55.0} & \textbf{62.2} \\
    \bottomrule
  \end{tabular}}
\end{minipage}
\end{table}

\begin{figure}[t]
    \centering
    \vspace{-5mm}
    \begin{minipage}[t]{0.48\textwidth}
        \centering
        \includegraphics[width=\linewidth]{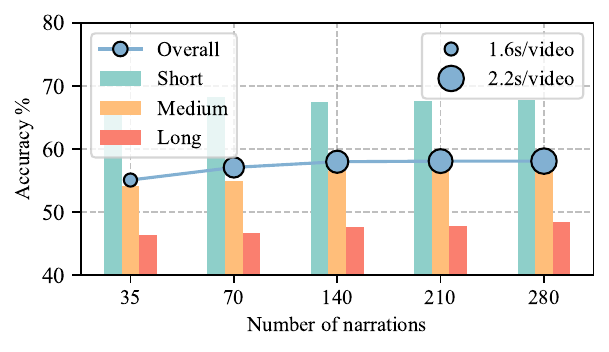}
        \vspace{-6mm}
        \caption{Effect of the number of inserted narratives. The average inference time per video increase from 1.6s to 2.2s with more narratives.}
        \label{fig:cap_num}
    \end{minipage}
    \hfill
    \begin{minipage}[t]{0.48\textwidth}
        \centering
        \includegraphics[width=\linewidth]{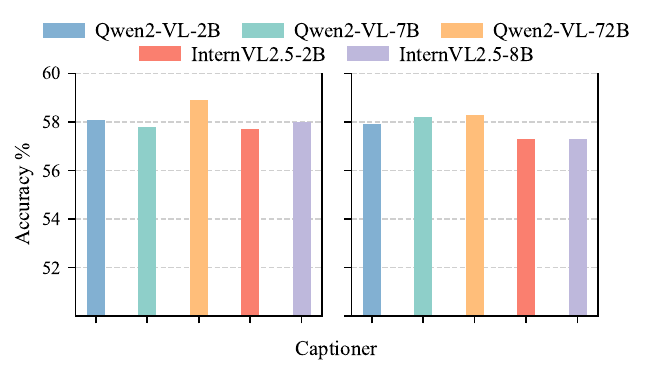}
        \vspace{-6mm}
        \caption{Impact of different captioners for generating narratives. Video-MME (sub.) results are for InternVL2-8B (left) and Qwen2-VL-7B (right).}
        \label{fig:captioner}
    \end{minipage}
    \vspace{-5mm}
\end{figure}

\vspace{-3mm}
\subsection{Ablation and Analysis}
\label{sec:ablation}

\textbf{KFC and Nar-KFC ablations.}
We report the ablation results of KFC and Nar-KFC components on the Video-MME (sub.) and MLVU benchmarks in Tab.~\ref{tab:ablation}.
Simply inserting narratives between uniformly sampled frames yields improvements of 2.9\% on Video-MME and 5.1\% on MLVU, indicating that adding narrative context, despite with frames not being query-specific, can effectively boost overall video understanding.
To retrieve query-relevant and diverse keyframes, our Greedy Search (GS) strategy achieves results comparable to the optimal Integer Quadratic Programming (IQP) method (55.0\% \textit{vs.} 55.1\% on Video-MME and 62.2\% \textit{vs.} 62.0\% on MLVU), while being significantly more efficient with $O(NK)$ complexity. Details of our IQP implementation and comparisons with GS are provided in Appendix \S\ref{sec:sup_iqp_gs}.
%3
Further ablations show that removing the query-relevance score $\mathcal{S}_{\mathrm{QR}}$ leads to a 3.2\% drop on Video-MME and 4.9\% on MLVU with greedy search.
This emphasizes that retrieving query-relevant frames is critical in long videoQA. Meanwhile, incorporating frame diversity $\mathcal{S}_{\mathrm{FD}}$ further stabilizes and enhances performance across benchmarks.
When threading all keyframes with interleaved narratives, Nar-KFC achieves the best overall results on all metrics, underscoring its solid effectiveness in representing long video contents.

\textbf{Component analysis of greedy search (GS). }
Starting from the vanilla GS, which iteratively selects the frame with the highest cumulative score relative to the already selected frames, we progressively incorporate several techniques (Tab.~\ref{tab:gs_module}) to enhance its effectiveness to a near-optimal solution:
(i) initialization with the frame most relevant to the query brings a modest yet consistent gain (from 52.3\%$\rightarrow$53.3\% on Video-MME, and 60.4\%$\rightarrow$61.0\% on MLVU);
(ii and iii) applying low-rank denoising %singular value decomposition (SVD)
and downsampling further improves performance by producing a more compact and less noisy score matrix $\mathbf{S}$;
and (iv) adding the final refinement step, KFC (GS) achieves the best results of 55.0\% on Video-MME and 62.2\% on MLVU. 
This highlights the cumulative benefit of combining compact frame representations, reduced redundancy, and an iterative selection mechanism. 
%
% Overall, these results validate the design of our KFC (GS) strategy as an effective and efficient solution for long video keyframe selection.

% \setlength{\columnsep}{1pt}%
\begin{wraptable}{r}{0.27\textwidth}
\centering
\vspace{-5mm}
\caption{Accuracies (\%) for using different frame extraction methods on Video-MME.}
\label{tab:kf_extraction}
% \vspace{-4.5mm}
\resizebox{0.27\textwidth}{!}{
  \begin{tabular}{@{}lc@{}}
\toprule
\multicolumn{2}{r}{V-MME$_\mathrm{(no\ sub. / sub.)}$} \\
\midrule
InternVL2 & 51.9 / 52.5 \\
\ + CLIP (top-K) & 47.7 / 50.0 \\
\ + SigLIP (top-K) & 47.3 / 51.0 \\
\ + BLIP-2 (top-K) & 47.8 / 50.9 \\
\ + TempGQA & 50.4 / 51.1 \\
\ + SeViLA & 52.2 / 53.7 \\
% + InternViT-6B & \\
\ + DPP & 52.2 / 53.5 \\
\ + AKS & 52.8 / 53.9 \\   % further checking
\ + KFC (Ours) & {\bf 53.1} / {\bf 55.0} \\
\bottomrule
\end{tabular}
  }
\vspace{-3.5mm}
\end{wraptable}

\textbf{Keyframe extraction baselines.}
We compare KFC with several keyframe extraction baselines in Tab.~\ref{tab:kf_extraction}, all utilizing the same InternVL2 backbone and 8 frames for a fair comparison. Implementation details are presented in Appendix \S\ref{sec:sup_kf_extract_baseline}.
Baseline methods that apply top-K frame-query matching using CLIP~\cite{radford2021CLIP}, SigLIP~\cite{zhai2023SigLIP}, or BLIP-2~\cite{li2023blip2} embeddings perform worse than uniform sampling, possibly due to the issue of keyframes being concentrated within a narrow temporal window.
For those first-localize-then-answer methods, i.e., TempGQA~\cite{xiao2024can} and SeViLA~\cite{yu2023selfchain}, performance heavily depends on the quality of segment localization, which tends to be less robust compared to KFC.
Recent proposed DPP~\cite{sun2025mdp3} and AKS~\cite{tang2025AKS} generally yield better results by incorporating video-level frame diversity. However, DPP relies on specially designed kernel functions, while AKS depends on handcrafted and heuristic sampling strategies.
Overall, our proposed KFC outperforms all the baselines, demonstrating clear superiority in subset frame selection and efficient long video understanding.

\textbf{Effect of narrative quantity.}
% \begin{figure}
%   \centering
%   % \fbox{\rule[-.5cm]{0cm}{4cm} \rule[-.5cm]{4cm}{0cm}}
%   \includegraphics[width=0.5\textwidth]{figure/caption_num_v2.pdf}
%   \caption{Effect of the number of inserted narratives.}
%   \label{fig:cap_num}
% \end{figure}
Fig.~\ref{fig:cap_num} shows the impact of the number of inserted narratives on Video-MME (sub.). We incrementally add narratives across 7 intervals between 8 keyframes.
The overall accuracy improves steadily from 55.1\% to 58.1\% as more narratives are available, with more performance gains on medium and long videos.
However, since adjacent frames often contain similar visual information, adding more narratives results in diminishing returns due to redundant descriptions. 
We thus use 210 narratives as the default, which takes $\sim$2.1 sec per video.

\textbf{Effect of narrative quality.}
% \begin{figure}
%   \centering
%   % \fbox{\rule[-.5cm]{0cm}{4cm} \rule[-.5cm]{4cm}{0cm}}
%   \includegraphics[width=0.6\textwidth]{figure/captioner.pdf}
%   \caption{Impact of different captioners. Results are shown on InternVL2-8B (left) and Qwen2-VL-7B (right).}
%   \label{fig:captioner}
% \end{figure}
Fig.~\ref{fig:captioner} presents the impact of different captioners on the quality of generated narratives and the resulting performance of Nar-KFC on the Video-MME (sub.). 
We evaluate five MLLMs of varying sizes and sources as captioners. 
Narratives extracted from the largest captioner, Qwen2-VL-72B, achieves the best accuracy, i.e., 58.9\% on InternVL2-8B and 58.3\% on Qwen2-VL-7B, highlighting the benefit of higher-quality narratives. 
Nevertheless, the overall performance gap across all captioners is small (less than 1\%). This suggests that keyframes play a dominant role in long video understanding, while captions serve as auxiliary and supportive context. 
% Given the trade-off between performance and efficiency, we adopt the lightweight Qwen2-VL-2B as the default captioner for other benchmarks.
We thus use the lightweight Qwen2-VL-2B as the default captioner for other benchmarks.

% \begin{table}[ht]
% \centering
% \label{tab:kf_extraction}
% \caption{Accuracies (\%) for using different frame extraction methods on Video-MME benchmark. }
% \begin{tabular}{lc}
% \toprule
%  & V-MME \\
% \midrule
% InternVL2 & 51.9 / 52.5 \\
% + CLIP & 47.7 / 50.0 \\
% + SigLIP & 47.3 / 51.0 \\
% + BLIP2 & 47.8 / 50.9 \\
% + TempGQA & \\
% + SeViLA & \\
% % + InternViT-6B & \\
% + DPP & 52.2 / 53.5 \\
% + AKS & 52.8 / 53.9 \\   % further checking
% + KFC (Ours) & 53.1 / 55.0 \\
% \bottomrule
% \end{tabular}
% \end{table}

\begin{table}[t] % ← 告诉 LaTeX 优先放在页面顶部
\vspace*{-1em}   % ← 减少默认浮动体顶部间距，可调节
\centering
\begin{minipage}{0.5\textwidth}
  \centering
  \captionof{table}{Analysis of video input components on Video-MME (no sub). Superscript numbers indicate the quantity.
  %V-MME denotes Video-MME (no sub.) performance. 
  Average time and tokens per video are reported.}
  \label{tab:pure_component}
  \setlength{\tabcolsep}{3pt}
  \resizebox{0.92\textwidth}{!}{
  \begin{tabular}{lccr}
    \toprule
    Components & V-MME & Time (s) & Token\#\\
    \midrule
    Narratives$^{210}$ &  51.1 & 0.98 &  4,725 \\
    Frames$^8$ (uniform) & 51.9 & 1.03 & 6,280 \\
    Frames$^8$ (KFC) & 53.1 & 1.31 & 6,280 \\
    \midrule    
    Interleave$^{8+210}$ (Nar-KFC) & 56.3 &  2.13 & 11,005 \\
    \bottomrule
  \end{tabular}}
\end{minipage}
% \hspace{0.06\textwidth} % 手动调整左右间距
\hfill
\begin{minipage}{0.44\textwidth}
  \centering
  \captionof{table}{Temporal structure analysis between narratives and keyframes on Video-MME (no sub).} % Results upon Video-MME (no sub.) are reported.}
  \label{tab:temporal}
  \vspace{2mm}
  \setlength{\tabcolsep}{3pt}
  \resizebox{0.95\textwidth}{!}{
  \begin{tabular}{lc}
    \toprule
    Temporal Structure & V-MME \\
    \midrule
    \{Narrative\}$\rightarrow$\{Keyframe\}$\rightarrow$\{Query\} & 55.5 \\
    \{Keyframe\}$\rightarrow$\{Narrative\}$\rightarrow$\{Query\} & 55.3 \\
    \midrule
    Interleave (Nar-KFC)$\rightarrow$\{Query\} & 56.3 \\
    \bottomrule
  \end{tabular}}
\end{minipage}
\vspace{-3mm}
\end{table}

\textbf{Efficiency and effectiveness between narratives and keyframes.}
We decompose Nar-KFC into standalone narratives and frames, and conduct thorough ablations in Tab.~\ref{tab:pure_component}.
Although translated from 210 frames, pure narratives perform worse than even 8 uniformly sampled frames (51.1\% \textit{vs.} 51.9\%), which reflects that substantial information is lost during the frame-to-caption conversion.
Nevertheless, narratives exhibit advantages in inference efficiency, requiring the shortest process time (0.98s) and the fewest tokens (4,725 per video).
Combining narratives with KFC-selected keyframes (Nar-KFC) achieves both the best accuracy and also maintains reasonable efficiency.
In addition, Tab.~\ref{tab:temporal} investigates the temporal structure between narratives and keyframes. Placing all keyframes either before or after the narratives degrades the performance by 0.8\% and 1.0\%, likely due to disrupted temporal sequences. 
In contrast, interleaving narratives and frames, as in Nar-KFC, yields superior results.
These findings further validate our primary goal: constructing temporally continuous representations for long video understanding.

\begin{figure}
  \centering
  % \fbox{\rule[-.5cm]{0cm}{3cm} \rule[-.5cm]{3cm}{0cm}}
  \includegraphics[width=1.0\textwidth]{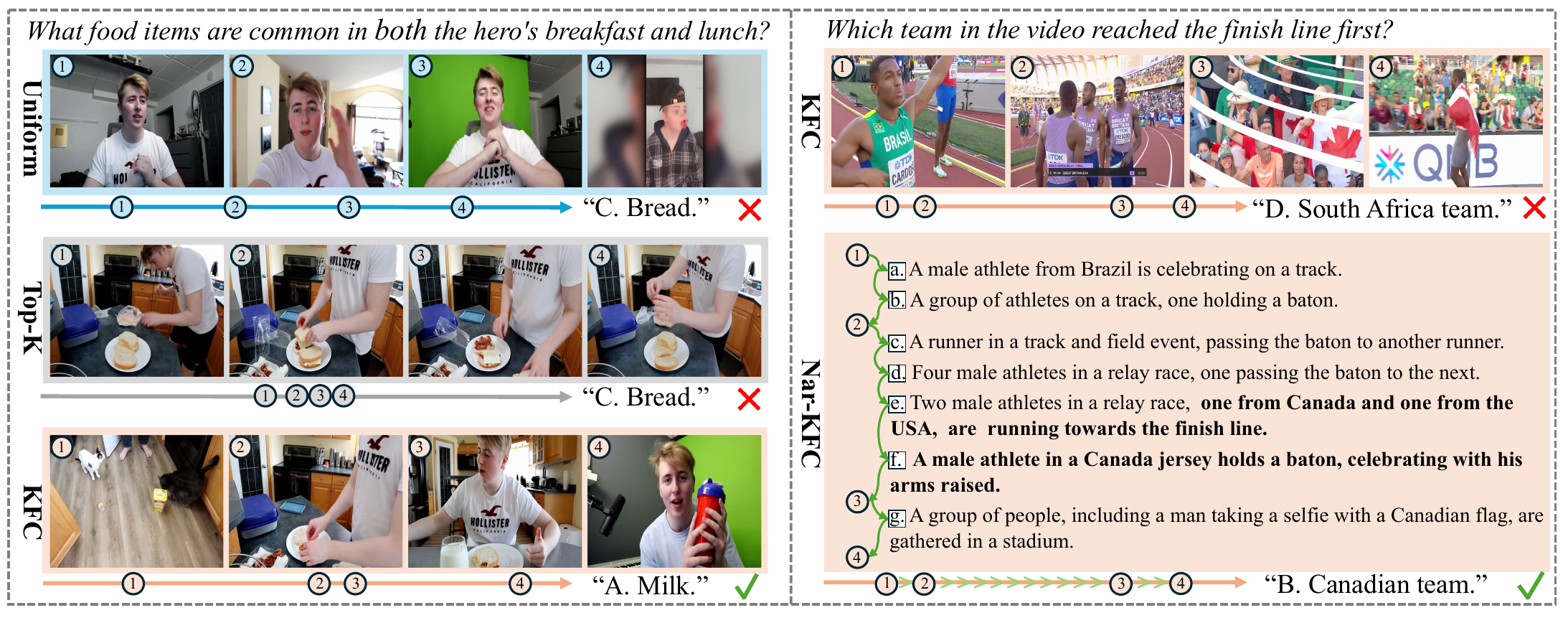}
  \vspace{-6mm}
  \caption{Qualitative results. (left) Comparison of frames selected by uniform sampling, top-K sampling, and our KFC. (right) Key narratives generated by Nar-KFC that lead to the correct answer.
  Both cases are from Video-MME dataset and use InternVL2 for inference. Zoom in for details.}
  \label{fig:vis}
  \vspace{-5mm}
\end{figure}

\vspace{-3mm}
\subsection{Qualitative Results}
\vspace{-2mm}

Fig.~\ref{fig:vis} presents two qualitative examples of our method. 
In the first example (left), uniform sampling struggles to capture relevant frames for the specific question, while top-k sampling yields similar frames concentrated within a narrow temporal window.
Neither approach successfully answers the question.
In contrast, our KFC effectively identifies frames that are both query-relevant and content-diverse, resulting in the correct answer.
In the second example (right), we demonstrate that Nar-KFC substantially improves reasoning in a complete relay race scenario by threading temporally interleaved keyframes with coherent narratives. This enables accurate inference of the final winner, whereas KFC fails due to limited number of frames.
More examples can be found in Appendix \S\ref{sec:sup_vis}.

%----------------------------------------------------------------------------
\vspace{-2.5mm}
\section{Conclusion}
\vspace{-2mm}
\label{sec:conclusion}
In this paper, we propose a keyframe capturing strategy (KFC) and a narrating keyframe method (Nar-KFC) to boost existing MLLMs for long video understanding, under the constraint of limited context length in language models. 
Our approach constructs long video representations that are query-relevant, content-diverse, and, importantly, temporally continuous, all achieved in a training-free manner. 
This significantly improves the performance of current MLLMs on widely-used long video benchmarks. Our findings strongly validate the potential of MLLMs as effective long video comprehenders. We discuss limitations and future works of Nar-KFC in Appendix \S\ref{sec:limtation}.

% \textbf{Limitations and future work.} 
% (1) \textit{Misalignment between MLLM training and our Nar-KFC testing.} Despite current MLLMs being able to process our interleaved inputs of keyframes and narratives, thanks to their instruction tuning step, they are not trained with such input formats. This may weaken their ability to fully understand the structure and relationships within our specialized long video representations. 
% A valuable future direction is to incorporate keyframe selection and narrative interleaving into the training of MLLMs, thereby aligning training and testing procedures for improved long video understanding.
% %
% (2) \textit{Coherence between keyframes and narratives.} The current narrative generation process relies on an off-the-shelf captioning model, which does not explicitly enforce temporal coherence across adjacent frames or clips. 
% A natural extension is to enhance narrative consistency by adopting a sliding-window captioning approach, such as the method proposed in ShareGPT4Video~\cite{chen2024sharegpt4video}, which we consider a valuable future direction.
% %
% (3) \textit{Difficulty in deploying Nar-KFC for real-time long video systems.} The current pipeline depends on frame-level CLIP embeddings and external captioning models, which can be time-consuming. This limits its applicability in real-time settings where a new long video must be processed and reasoned over promptly. Enabling Nar-KFC to operate efficiently in real-time remains an open and challenging problem.

{
\small
\bibliographystyle{plain}
\bibliography{ref}
}

\clearpage
\appendix
% \section*{\centering \Large \textbf{APPENDIX}}
% --------- 美观的补充材料页标题 ----------
\begin{center}
    {\LARGE \textbf{Threading \textit{Keyframe} with \textit{Narratives}:}}\\[0.5em]
    {\LARGE \textbf{MLLM as Strong Long Video Comprehender}}\\[1.2em]
    {\Large {Supplemental Material}}\\[1.0em]
    % 如果需要作者/机构信息，可加下方这行
    % \vspace{1em}
    % {\normalsize Author Names and Affiliations Here}
\end{center}
\vspace{1.5em}
% --------------------------------------------

\renewcommand{\thesection}{\Alph{section}} % A, B, C...
\setcounter{table}{0}
\setcounter{figure}{0}
\renewcommand{\thetable}{\Alph{section}.\arabic{table}}   % Table A.1
\renewcommand{\thefigure}{\Alph{section}.\arabic{figure}} % Figure A.1

\section{Limitations and Future Work}
\label{sec:limtation}
We discuss limitations and possible extensions of Nar-KFC.
(1) \textit{Misalignment between MLLM training and Nar-KFC testing.} Despite current MLLMs being able to process our interleaved inputs of keyframes and narratives, thanks to their instruction tuning step, they are not trained with such input formats. This may weaken their ability to fully understand the structure and relationships within our specialized long video representations. 
A valuable future direction is to incorporate keyframe selection and narrative interleaving into the training of MLLMs, thereby aligning training and testing procedures for improved long video understanding.
(2) \textit{Coherence between keyframes and narratives.} The current narrative generation process relies on an off-the-shelf captioning model, which does not explicitly enforce temporal coherence across adjacent frames or clips. 
A natural extension is to enhance narrative consistency by adopting a sliding-window captioning approach, such as the method proposed in ShareGPT4Video~\cite{chen2024sharegpt4video}, which we consider a valuable future direction.
(3) \textit{Difficulty in deploying Nar-KFC for real-time long video systems.} The current pipeline depends on frame-level CLIP embeddings and external captioning models, which can be time-consuming. This limits its applicability in real-time settings where a new long video must be processed and reasoned over promptly. Enabling Nar-KFC to operate efficiently in real-time remains an open and challenging problem.

\section{Broader Impacts}
\label{sec:impact}

Effective and efficient long video understanding is a critical task, especially as Internet video streams often last tens of minutes or even hours.
We expect that the proposed keyframe selection and narration methods will benefit society by enabling MLLMs to comprehend long videos more accurately and efficiently.
However, it is essential to ensure that the narratives generated by specific models remain free from harmful or unrelated content.

\section{Main Results Supplementary}
\label{sec:sup_main_results}
% We supplement main results of earlier works in Sec.~\ref{sec:sup_full_sota}, the performance of our KFC and Nar-KFC on additional EgoSchema and NExTQA benchmarks in Sec.~\ref{sec:sup_sota}, and detailed analysis on MLUV benchmark in Sec.~\ref{sec:sup_mlvu_analysis}.
We provide supplementary results to the main experiments:
Sec.~\ref{sec:sup_full_sota} covers earlier works, 
Sec.~\ref{sec:sup_sota} presents the performance of KFC and Nar-KFC on additional EgoSchema and NExTQA benchmarks, and 
Sec.~\ref{sec:sup_mlvu_analysis} offers detailed analysis on the MLVU benchmark.

\subsection{Comprehensive Comparisons with Previous Methods.}
\label{sec:sup_full_sota}

\begin{table}
  \caption{Comprehensive comparisons with previous VideoLLMs/MLLMs on three common long-video benchmarks: Video-MME, LVB$_{val}$, and MLVU.
  For Video-MME, we report performance with two standard settings: without subtitles (no sub.) and with subtitles (sub.).
  LVB denotes the LongVideoBench validation set, with results for (15m, 60m] long videos shown in parentheses.
  Methods that use significantly more frames and larger-sized LLM are marked in \textcolor{gray}{gray}.
  The reported results are accuracy percentage.
  %Accuracy sign \% is omitted for clarity.
  }
  \label{tab:sup_full_sota}
  \centering
  \setlength{\tabcolsep}{5pt}  % 缩小列间距
  \resizebox{\textwidth}{!}{
  \begin{tabular}{lccccccc}
    \toprule
    \multirow{2}{*}{Model} & \multirow{2}{*}{Size} & \multicolumn{4} {c}{Video-MME$_{\mathrm{(no\ sub.\ /\ sub.)}}$} & \multirow{2}{*}{LVB} & \multirow{2}{*}{MLVU}\\
    \cmidrule(r){3-6}
    & & Short    & Medium     & Long & Overall$_{\sim 17m}$ & $_{\sim 12m}$  & $_{\sim 12m}$ \\
    \hline
    Video-LLaVA~\cite{lin2024video_llava} & 7B & 45.3 / 46.1 &  38.0 / 40.7 & 36.2 / 38.1 & 39.9 / 41.6 & 39.1 & 47.3 \\
    Qwen-VL-Chat~\cite{bai2023qwen} & 7B & 46.9 / 47.3 & 38.7 / 40.4 & 37.8 / 37.9 & 41.1 / 41.9 & -& - \\
    ST-LLM~\cite{liu2024st-llm} & 7B & 45.7 / 48.4 &  36.8 / 41.4 & 31.3 / 36.9 &  37.9 / 42.3 & - & - \\
    VideoChat2~\cite{li2023videochat} & 7B & 48.3 / 52.8 & 37.0 / 39.4 & 33.2 / 39.2 & 39.5 / 43.8 & 39.3 & 44.5 \\
    ShareGPT4Video~\cite{chen2024sharegpt4video} & 8B & 48.3 /\pos & 36.3 /\pos & 35.0 /\pos & 39.9 /\pos & 41.8 & 46.4 \\
    Chat-UniVi-V1.5~\cite{jin2024chat-univi} & 7B &  45.7 / 51.2 & 40.3 / 44.6 & 35.8 / 41.8  & 40.6 / 45.9 & - & - \\
    VideoLLaMA2~\cite{cheng2024videollama} & 7B & 56.0 /\pos & 45.4 /\pos & 42.1 /\pos & 47.9 /\pos & - & - \\
    VILA~\cite{lin2024vila} & 8B & 57.8 / 61.6 & 44.3 / 46.2 & 40.3 / 42.1 & 47.5 / 50.0 & - & 46.3 \\
    LLaVA-NeXT-QW2~\cite{liu2024llavanext} & 7B & 58.0 /\pos  & 47.0 /\pos & 43.4 /\pos & 49.5 /\pos & {-} & {-} \\
    MiniCPM-V2.6~\cite{yao2024minicpm} & 7B  & 61.1 / 63.8 & 50.3 / 50.2 & 46.4 / 45.4 & 52.6 / 53.1 & 51.2 & 55.4 \\
    LongVU~\cite{shen2024longvu} & 7B & 64.7 /\pos & 58.2 /\pos & 59.5 /\pos & 60.6 /\pos & - & 65.4 \\
    Frame-Voyager~\cite{yuframe-voyager} & 8B & 67.3 /\pos & 56.3 /\pos & 48.9 /\pos & 57.5 /\pos & - & 65.6 \\
    \cg LongVILA$^{\mathrm{256frm}}$~\cite{chen2024longvila} & \cg 8B & \cg 61.8 /\pos & \cg 49.7 /\pos & \cg 39.7 /\pos & \cg 50.5 /\pos & \cg - & \cg -\\
    \cg Video-XL$^{\mathrm{256frm}}$~\cite{shu2024videoxl} & \cg 7B & \cg 64.0 / 67.4 & \cg 53.2 / 60.7 &\cg 49.2 / 54.9 &\cg 55.5 / 61.0 &\cg 50.7 &\cg 64.9 \\
    \cg LLaVA-NeXT-Video~\cite{zhang2024llavanextvideo} &\cg 34B &\cg  61.7 / 65.1 &\cg 50.1 / 52.2 &\cg 44.3 / 47.2 &\cg 52.0 / 54.9 &\cg 50.5 &\cg 58.8 \\
    \cg VILA~\cite{lin2024vila} &\cg 34B &\cg 70.3 / 73.1 &\cg 58.3 / 62.7 &\cg 51.2 / 55.7 &\cg 58.3 / 61.6 &\cg - &\cg 57.8 \\
    \hline
    InternVL2~\cite{chen2024internvl} & 8B & 62.1 / 63.9 & 48.2 / 48.7 & 45.2 / 44.9 & 51.9 / 52.5 & 52.3 (45.2) & 54.3 \\
    \rowcolor{lighterblue}
    \quad \textbf{+ KFC} & 8B & 64.3 / 65.4 & 49.6 / 52.3 & 46.1 / 47.3 & 53.1 / 55.0 & 53.3 (47.2) & 62.2 \\
    \rowcolor{lightblue}
    \quad \textbf{+ Nar-KFC} & 8B & 67.2 / 67.7 & 54.7 / 57.9 & 47.1 / 48.9 & \textbf{56.3} / \textbf{58.1} & \textbf{53.9} (\textbf{48.8}) & \textbf{64.4} \\
    \hline
    Qwen2-VL~\cite{wang2024qwen2vl} & 7B & 65.7 / 66.9 & 52.8 / 53.0 & 46.7 / 48.6 & 55.0 / 56.1 & 53.4 (45.0) & 59.6\\
    \rowcolor{lighterblue}
    \quad \textbf{+ KFC} & 7B & 68.2 / 69.7 & 53.3 / 54.9 & 48.4 / 50.2 & \textbf{56.7} / \textbf{58.3} & \textbf{54.6} (\textbf{47.9}) & 65.9 \\
    \rowcolor{lightblue}
    \quad \textbf{+ Nar-KFC} & 7B & 68.8 / 69.3 & 53.4 / 55.3 & 48.0 / 49.0 & \textbf{56.7} / 57.9 & 53.6 (46.3) & \textbf{68.5} \\
    \hline
    LLaVA-OneVision~\cite{li2024llava_ov} & 7B  & 65.2 / 67.1 & 51.7 / 54.4 & 45.1 / 46.1 & 53.3 / 55.9 & 54.5 (45.7) & 58.5 \\
    \quad + BOLT~\cite{liu2025bolt} & 7B & 66.8 /\pos & 54.2 /\pos & 47.3 /\pos & 56.1 /\pos & 55.6 (\quad -\ \ ) & 63.4 \\
    \rowcolor{lighterblue}
    \quad \textbf{+ KFC} & 7B & 66.4 / 69.1 & 52.9 / 56.8 & 46.8 / 48.8 & 55.4 / 58.2 & 55.6 (47.3) & 65.0 \\
    \rowcolor{lightblue}
    \quad \textbf{+ Nar-KFC} & 7B & 67.2 / 68.6 & 57.1 / 59.8 & 49.1 / 51.0 & \textbf{57.8} / \textbf{59.8} & \textbf{56.5} (\textbf{48.2}) & \textbf{66.2} \\
    \hline
    LLaVA-Video~\cite{zhang2024llava_video} & 7B & 67.2 / 69.4 & 53.2 / 53.4 & 47.2 / 47.3 & 55.9 / 56.7 & 54.2 (46.5) & 60.5 \\
    \rowcolor{lighterblue}
    \quad \textbf{+ KFC} & 7B & 68.3 / 70.0 & 55.1 / 57.4 & 49.4 / 51.6 & 57.6 / 59.7 & 56.5 (49.3) & 66.9 \\
    \rowcolor{lightblue}
    \quad \textbf{+ Nar-KFC} & 7B & 71.2 / 72.7 & 61.4 / 62.3 & 52.0 / 53.9 & \textbf{61.6} / \textbf{63.0} & \textbf{57.7} (\textbf{50.2}) & \textbf{67.7} \\
    \bottomrule
  \end{tabular}
  }
\end{table}
VideoLLMs for video understanding have become a popular research area in recent years. However, directly applying previous VideoLLMs to long videos, such as Video-MME, LongVideoBench, and MLVU, often leads to unsatisfactory performance. 
To provide a more comprehensive comparison, as an extension to the main paper in Tab.~\ref{tab:sota1}, we also include the performance of earlier works, such as  Video-LLaVA, Qwen-VL-Chat, ST-LLM, VideoChat2, ShareGPT4Video, Chat-UniVi-V1.5, and VideoLLaMA2, in Tab.~\ref{tab:sup_full_sota}.

Notably, we compare the performance of the very recent BOLT~\cite{liu2025bolt} method, using the same backbone, i.e., LLaVA-OneVision, and the same 8 frames. 
This comparison can be seen as an additional keyframe extraction baseline in the main paper Tab.~\ref{tab:kf_extraction}.
BOLT adopts inverse transform sampling for long video keyframe extractions.
Generally, it performs slightly better than KFC on Video-MME (56.1\% \textit{vs.} 55.4\%), while it is worse than KFC on MLVU (63.4\% \textit{vs.} 65.0\%). These results further confirm the effectiveness of KFC.
Additionally, incorporating interleaved narratives (Nar-KFC) yields further solid improvements.

\subsection{Results on more Benchmarks}
\label{sec:sup_sota}
\vspace{-3mm}
\begin{table}[ht]
\centering
\caption{Results on EgoSchema and NExTQA benchmarks. Accuracy sign \% is omitted for clarity.}
\label{tab:sup_ego_nextqa}
\begin{tabular}{lcc}
\toprule
\multirow{2}{*}{Model} & EgoSchema & \multirow{2}{*}{NExTQA} \\
& \textit{Subset} & \\
\midrule
InternVL2-8B & 59.8 & 77.8 \\
\rowcolor{lighterblue}
\ + KFC & 58.6 & 77.8 \\
\rowcolor{lightblue}
\ + Nar-KFC & \textbf{64.0} & \textbf{78.1} \\
\midrule
Qwen2-VL-7B & 60.8 & 77.3 \\
\rowcolor{lighterblue}
\ + KFC & 63.2 & 76.6 \\
\rowcolor{lightblue}
\ + Nar-KFC & \textbf{65.8} & \textbf{77.6} \\
\bottomrule
\end{tabular}
\end{table}

We further report performance of our KFC and Nar-KFC on two relatively shorter video benchmarks, i.e., EgoSchema (Subset) and NExTQA, in Tab.~\ref{tab:sup_ego_nextqa}.

\textit{EgoSchema}~\cite{mangalam2023egoschema}. EgoSchema is a benchmark derived from 5,000 egocentric videos, capturing a first-person perspective of humans engaged in a wide range of daily activities, each lasting approximately 3 min. Here, we use its subset of 500 questions with publicly available labels. 

\textit{NExTQA}~\cite{xiao2021nextqa}. Following standard practices, we use the validation set of NExTQA for evaluation. This set contains 570 videos and 5,000 multiple-choice questions, with an average of 44 sec.

Unlike the long video datasets discussed in the main paper, our keyframe selection strategy (i.e., KFC) may underperfom compared to uniform sampling when applied to shorter videos. For example, InternVL2-8B yields 58.6\% accuracy on EgoSchema and Qwen2-VL-7B achieves 76.6\% on NExTQA when using KFC. This performance drop is primarily due to KFC disrupting the temporal consistency of frame sequences, which is particularly important for short video understanding.
Nevertheless, supplementing with non-keyframe narratives (Nar-KFC) leads to consistent performance improvements even on these shorter benchmarks. The gains are especially evident on EgoSchema, while the improvement on NExTQA is more limited, likely due to its relatively short average video length of approximately 44 sec.

\subsection{Detailed Analysis on MLVU Categories}
\label{sec:sup_mlvu_analysis}

\begin{figure}[ht]
  \centering
  % \fbox{\rule[-.5cm]{0cm}{4cm} \rule[-.5cm]{4cm}{0cm}}
  \includegraphics[width=1.0\textwidth]{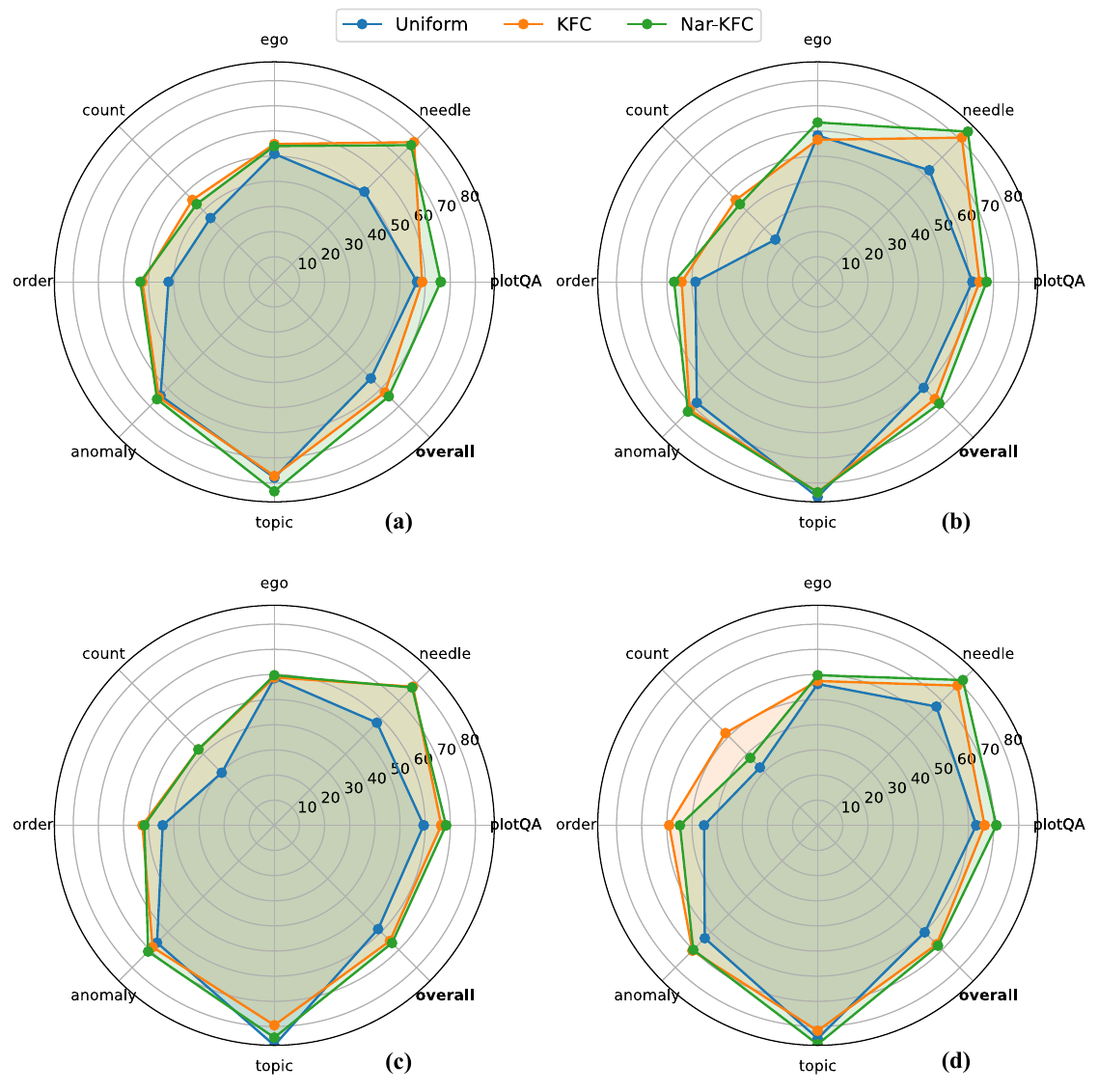}
  \caption{Performance comparison across specific categories of the MLVU benchmark. Results are shown for (a) InternVL2-8B, (b) Qwen2-VL-7B, (c) LLaVA-OneVision-7B, and (d) LLaVA-Video-7B, evaluated using three keyframe selection strategies: Uniform, KFC, and Nar-KFC.}
  \label{fig:sup_mlvu}
\end{figure}

In Fig.\ref{fig:sup_mlvu}, we provide a detailed comparison of performance across specific categories in the MLVU benchmark as a supplement to the main paper Tab.~\ref{tab:sota1}. 
Compared to uniform sampling, the overall performance improvement introduced by KFC across all four models is primarily attributed to its superior accuracy in the \textbf{needle} and \textbf{count} categories.
The \textit{needle} task involves questions based on rare or unusual frames sourced from external videos, which are more likely to be captured by our query-relevance-based sampling strategy. In contrast, such frames are often missed by uniform sampling.
A similar challenge arises in the \textit{count} task, where correct answers rely on retrieving specific frames first in order to support accurate object/crowd/event counting.

On the other hand, our Nar-KFC approach generally achieves the best performance on \textbf{plotQA} and \textbf{topic} tasks. This advantage stems from its ability to preserve temporal continuity, which is often lacking in KFC-optimized keyframes that are temporally sparse and discontinuous. Such discontinuity hinders the model's ability to comprehend holistic video contents. 
For instance, KFC performs the worst on the \textit{topic} task when inferenced with LLaVA-OneVision (c) and LLaVA-Video (d), even underperforming the uniform sampling baseline.
In contrast, Nar-KFC addresses this issue through a narrative threading strategy, which maintains continuity by supplementing keyframes with coherent non-keyframe descriptions. This strategy significantly enhances the model's understanding of overall video plots and topics.

\section{Additional Ablation Results}
\label{sec:sup_ablation}

\subsection{A Symmetrical Formulation of Original Objective and Analysis.}
\label{sec:sup_symmtric}

\textbf{Objective Revisiting.} In the main paper Sec.~\ref{sec:method}, we formulate the keyframe selection task as a \textit{graph} problem and model it using integer quadratic programming (IQP) (\ref{eq:iqp}). 
However, the constructed score matrix (\ref{eq:score}) is asymmetric, as it only accounts for the query relevance of the $i$-th frame and the diversity between the $i$-th and $j$-th frames, while neglecting the query relevance of the $j$-th frame. This asymmetry introduces a minor discrepancy compared to the standard subgraph selection procedure. 
We illustrate this discrepancy with an example.

\textbf{Example.} Suppose we aim to retrieve 3 keyframes from 5 frames, and the optimal selection is given by $\mathbf{x}=[1,1,1,0,0]^T$, indicating that first three frames are selected. The score matrix $\mathbf{S}$ is defined as:
\begin{equation}
     \mathbf{S}_{i,j}= \mathcal{S}(i,j) = S_{\mathrm{QR}}(i) + S_{\mathrm{FD}}(i, j) = 
     \begin{bmatrix}
     0 & a_{12} & a_{13} & a_{14} & a_{15} \\
     0 & 0 & a_{23} & a_{24} & a_{25} \\
     0 & 0 & 0 & a_{34} & a_{35} \\
     0 & 0 & 0 & 0 & a_{45} \\
     0 & 0 & 0 & 0 & 0 \\
     \end{bmatrix}.
\end{equation}
where $a_{i,j}$ denotes the score term for $i<j$ (i.e., only the upper trangular part of $\mathbf{S}$ is considered). According to (\ref{eq:iqp}), the maximum sum score (the total edge weight of the subgraph) should be:
\begin{equation}
\begin{aligned}
    \mathbf{x}^T\mathbf{S}\mathbf{x} &= [1,1,1,0,0]
    \begin{bmatrix}
     0 & a_{12} & a_{13} & a_{14} & a_{15} \\
     0 & 0 & a_{23} & a_{24} & a_{25} \\
     0 & 0 & 0 & a_{34} & a_{35} \\
     0 & 0 & 0 & 0 & a_{45} \\
     0 & 0 & 0 & 0 & 0 \\
    \end{bmatrix}
    [1,1,1,0,0]^T \\
    &= [1,1,1,0,0][a_{12}+a_{13}, a_{23}, 0, 0, 0]^T\\
    &= a_{12}+a_{13}+a_{23} \\
    &= S_{\mathrm{QR}}(1)+S_{\mathrm{FD}}(1,2) +
       S_{\mathrm{QR}}(1)+S_{\mathrm{FD}}(1,3) +
       S_{\mathrm{QR}}(2)+S_{\mathrm{FD}}(2,3).
\end{aligned}
\end{equation}
From this computation, we know that the query relevance of the first frame is counted twice, while that of the last selected frame (3$^{rd}$) is not counted at all, as there are no subsequent frames after it. 
This \textit{discrepancy} shows the deviation from the standard graph-based subgraph selection formulation.

\textbf{Symmetric Score Matrix.}
To mitigate this discrepancy and align the keyframe selection process with a standard graph problem, we reconstruct the original score matrix $\mathbf{S}$ to be symmetric by incorporating the query relevance of the $j$-th frame, defined as:
\begin{equation}
\label{eq:new_score}
    \mathbf{S}_{i,j}=S(i,j)=  S_{\mathrm{QR}}(i) + 2S_{\mathrm{FD}}(i, j) + S_{\mathrm{QR}}(j).
\end{equation}

\textbf{Experimental Results and Analysis.}
\begin{table}[ht]
\centering
\caption{Impact of whether replacing score matrix to its symmetric counterpart. Results are reported on the Video-MME (sub.) benchmark using InternVL2-8B model. The search node number is 40k for solving IQP.}
\label{tab:symmtric_score}
\begin{tabular}{lc|cccc}
\toprule
\multirow{2}{*}{Setting}& \multirow{2}{*}{Strategy} & \multicolumn{4}{c}{Video-MME (sub.)} \\
 & & Short & Medium & Long & Overall \\
\midrule
asymmetric $\mathbf{S}$ (\ref{eq:score}) & \multirow{2}{*}{IQP} & 65.9 & 52.9 & 46.4 & \textbf{55.1} \\
symmetric $\mathbf{S}$ (\ref{eq:new_score}) & & 66.1 & 50.1 & 46.1 & 54.1 \\
\midrule
asymmetric $\mathbf{S}$ (\ref{eq:score}) & \multirow{2}{*}{GS} & 65.4 & 52.3 & 47.3 & 55.0 \\
symmetric $\mathbf{S}$ (\ref{eq:new_score}) & & 65.7 & 52.6 & 47.2 & \textbf{55.1} \\
\bottomrule
\end{tabular}
\vspace{-4mm}
\end{table}
Compared with the symmetric $\mathbf{S}$ in (\ref{eq:new_score}), our original asymmetric matrix involves fewer terms with reducing size (only the upper triangular part is calculated), which leads to faster inference.
Tab.~\ref{tab:symmtric_score} presents additional experimental results for replacing the original score matrix $\mathbf{S}$ with its symmetric counterpart. 
Modifying $\mathbf{S}$ to be symmetric -- thus aligning the formulation with a standard graph problem -- 
results in a 1\% performance drop when using the IQP solver. 
This result supports the benefit of assigning higher weights to the initially selected frame at the beginning. 
Since the first keyframe is heuristically selected based on query relevance, this modification has negligible impact when using the GS strategy. We thus adopt the asymmetric score matrix defined in (\ref{eq:score}) for the remainder of our process.
% has minimal impact on performance. The slight decrease in performance (1\%) observed with IQP is likely due to instability in solving the IQP, which is further discussed in Sec.~\ref{sec:sup_iqp_gs}.
%
% In conclusion, we provide a comprehensive analysis and discussion that supports the main formulation in the main paper. The supplementary experiment reinforces the effectiveness of modeling keyframe selection as a graph problem.

%
\subsection{Integer Quadratic Programming (IQP) \textit{vs.} Greedy Search (GS)}
\label{sec:sup_iqp_gs}

\begin{table}[ht]
\centering
\caption{Impact of expanding the IQP search space on performance and efficiency. Results are reported on the Video-MME (sub.) benchmark using InternVL2-8B model, with average computational time per video (in seconds) evaluated on a single NVIDIA A100 GPU. }
\label{tab:sup_IQP}
\begin{tabular}{c|c|cccc|c}
\toprule
\multirow{2}{*}{Setting} & \multirow{2}{*}{Nodes\#} & \multicolumn{4}{c|}{Video-MME (sub.)} & \multirow{2}{*}{Time (s)} \\
&  & Short & Medium & Long & Overall & \\
\midrule
Uniform & - & 63.9 & 48.7 & 44.9 & 52.5 & 1.03 \\
GS & - & 65.4 & 52.3 & 47.3 & 55.0 & 1.31 \\
\midrule
& 5k & 64.2 & 52.6 & 46.7 & 54.5 & 3.91 \\
& 10k & 64.4 & 52.6 & 45.8 & 55.0 & 4.81 \\
IQP & 20k & 64.3 & 52.3 & \textbf{47.9} & 54.9 & 6.23 \\
& 30k & 65.6 & 52.6 & 46.2 & 54.8 & 8.01 \\
& 40k & \textbf{65.9} & \textbf{52.9} & 46.4 & \textbf{55.1} & 9.26 \\
\midrule
\multirow{5}{*}{\makecell{IQP\\(GS init)}} 
  & \cellcolor{lightgray}5k  & \cellcolor{lightgray}64.3 & \cellcolor{lightgray}52.0 & \cellcolor{lightgray}48.0 & \cellcolor{lightgray}54.7 & \cellcolor{lightgray}5.22 \\
  & \cellcolor{lightgray}10k & \cellcolor{lightgray}65.1 & \cellcolor{lightgray}51.9 & \cellcolor{lightgray}47.3 & \cellcolor{lightgray}54.5 & \cellcolor{lightgray}6.12 \\
  & \cellcolor{lightgray}20k & \cellcolor{lightgray}65.1 & \cellcolor{lightgray}52.3 & \cellcolor{lightgray}47.5 & \cellcolor{lightgray}54.9 & \cellcolor{lightgray}7.54 \\
  & \cellcolor{lightgray}30k & \cellcolor{lightgray}65.3 & \cellcolor{lightgray}52.3 & \cellcolor{lightgray}45.7 & \cellcolor{lightgray}54.4 & \cellcolor{lightgray}9.32 \\
  & \cellcolor{lightgray}40k & \cellcolor{lightgray}65.8 & \cellcolor{lightgray}51.4 & \cellcolor{lightgray}46.0 & \cellcolor{lightgray}54.4 & \cellcolor{lightgray}10.57 \\
\bottomrule
\end{tabular}
\end{table}

We implement the Integer Quadratic Programming (IQP) algorithm using CPLEX and set a maximum number of search nodes to obtain the optimal set of keyframe indices within a limited time.
The corresponding IQP results are reported in Tab.~\ref{tab:sup_IQP}.
As the search space increases from 5k to 40k nodes, performance on short videos gradually improves from 64.2\% to 65.9\%, which validates the effectiveness of modeling keyframe selection as an IQP problem.
However, this improvement does not hold for long videos, where performance becomes unstable as the search space expands. We speculate that this is because even 40k nodes are still insufficient to cover the full solution space for long videos. For instance, in a 15-minute video (900 frames at 1 fps), selecting 8 keyframes results in approximately $C(900, 8)\simeq 2.5\times 10^{18}$, i.e., roughly 2.5 quintillion possible combinations. This vast search space far exceeds what can be practically explored with a node limit of 40k, let alone for videos that span several hours.

We also attempt to initialize the IQP search with greedy searched results, which are highlighted in gray in Tab.~\ref{tab:sup_IQP}, in hopes of better guiding the IQP solving process. 
Experimental results indicate that this initialization strategy does not lead to further improvements in IQP performance, likely due to the search space remaining too large to be effectively navigated. 
Therefore, we adopt a customized greedy search (GS) strategy as a practical and robust alternative to the IQP algorithm.

\subsection{Ablations on Refinement Window Size \textit{k}}
\label{sec:sup_neighbor_k}

\begin{table}
\centering
\caption{Impact of refinement window size $k$ in our customized Greedy Seach (GS) algorithm.}
\label{tab:sup_neighbor_k}
\begin{tabular}{l|cccc}
\toprule
\multirow{2}{*}{Window Size $k$}& \multicolumn{4}{c}{Video-MME (sub.)} \\
& Short & Meidum & Long & Overall \\
\midrule
0 (w/o refine) & 65.0 & 51.2 & \textbf{47.8} & 54.7 \\
\midrule
1 & 65.1 & 51.4 & 47.4 & 54.7 \\
\rowcolor{lightgray}
2 & \textbf{65.4} & 52.3 & 47.3 & \textbf{55.0} \\
4 & 64.6 & \textbf{53.1} & 45.7 & 54.4 \\
8 & 64.2 & 50.3 & 43.8 & 52.8 \\
\bottomrule
\end{tabular}
\vspace{-4mm}
\end{table}

We analyze the impact of the neighbor window size $k$ in the final Greedy Search (GS) refinement step. As shown in Tab.~\ref{tab:sup_neighbor_k}, setting $k=0$ corresponds to using the GS strategy without any refinement. 
When $k=2$, which means examining a total of four frames, two before and two after the selected keyframe, the model achieves the best overall performance. This highlights the effectiveness of the refinement step as a robust strategy to complement prior SVD and downsampling operations.
However, increasing the window size further (e.g., $k=4$ or $k=8$) results in performance degradation. This is likely due to the disruption of holistic keyframe combinations constructed by the greedy search, as excessive frame examination may introduce noise or redundancy.

\subsection{Implementation Details of Frame Extraction Baselines in Tab.~\ref{tab:kf_extraction}}
\label{sec:sup_kf_extract_baseline}

For CLIP\footnote{https://huggingface.co/openai/clip-vit-large-patch14-336}~\cite{radford2021CLIP}, SigLIP\footnote{https://huggingface.co/google/siglip-so400m-patch14-384}~\cite{zhai2023SigLIP}, and BLIP-2\footnote{https://huggingface.co/Salesforce/blip2-opt-2.7b}~\cite{li2023blip2}, we directly rank and select the top-K candidate keyframes based on their frame-query cosine similarity logits.
For TempGQA~\cite{xiao2024can}, we follow the official code\footnote{https://github.com/doc-doc/NExT-GQA/tree/main/code/TempGQA} to first select a segment based on the question, and then uniformly sample frames from the selected segment to generate the answer.
For SeViLA~\cite{yu2023selfchain}, we use its trained localizer\footnote{https://github.com/Yui010206/SeViLA?tab=readme-ov-file} to select the $K$ keyframes as input, while maintaining the original hyperparameter settings.
As for DPP (Determinantal Point Process) selection~\cite{sun2025mdp3}, since the official code is unavailable, we reimplement the DPP algorithm by defining its kernel matrix as $\mathcal{S}(i, q)\mathcal{S}(j, q)[1-\mathcal{S}(i, j)]$, where the first two terms represent the similarity of frames $i, j$ to the query $q$, and the last term encourages frame diversity between frame $i$ and frame $j$. $\mathcal{S}$ denotes the cosine similarity operation.
For AKS~\cite{tang2025AKS}, we select keryframes based on the frame scores provided in the official repository\footnote{https://github.com/ncTimTang/AKS}.

\subsection{Incorporating Full Video-level Narratives.}
\label{sec:sup_all_narrative}

\begin{table}
\centering
\caption{Impact of incorporating full video-level narratives. These narratives include segments that appear before the first keyframe and after the last keyframe. $^*$ indicates that only narratives \textit{between keyframes} are utilized in Nar-KFC.}
\label{tab:all_nar}
\begin{tabular}{l|cccc}
\toprule
\multirow{2}{*}{Setting}& \multicolumn{4}{c}{Video-MME (no sub. / sub.)} \\
& Short & Meidum & Long & Overall \\
\midrule
Full-Narrative & 66.3 / 66.9 & 56.3 / 58.0 & 46.7 / 47.3 & \textbf{56.4} / 57.4 \\
Nar-KFC$^*$ & 67.2 / 67.7 & 54.7 / 57.9 & 47.1 / 48.9  & 56.3 / \textbf{58.1} \\
\bottomrule
\end{tabular}
\vspace{-4mm}
\end{table}

Our default Nar-KFC  configuration (see main paper Sec.~\ref{subsec:nar-kfc}) only uses narratives that appear between the first and the last keyframe, discarding those that occur at the beginning or end of the video. 
Here, we analyze the effect of incorporating full video-level narratives, as shown in Tab.~\ref{tab:all_nar}, while keeping the total number of inserted narratives fixed at 210.
The results suggest that including these additional narratives has minimal impact on overall video understanding. This finding further supports our primary conclusion: keyframes play a dominant role in long-form VideoQA, while narratives mainly serve as auxiliary context.

\vspace{-2mm}
\section{Additional Qualitative Examples}
\label{sec:sup_vis}
\begin{figure}[ht]
  \centering
  % \fbox{\rule[-.5cm]{0cm}{4cm} \rule[-.5cm]{4cm}{0cm}}
  \includegraphics[width=0.95\textwidth]{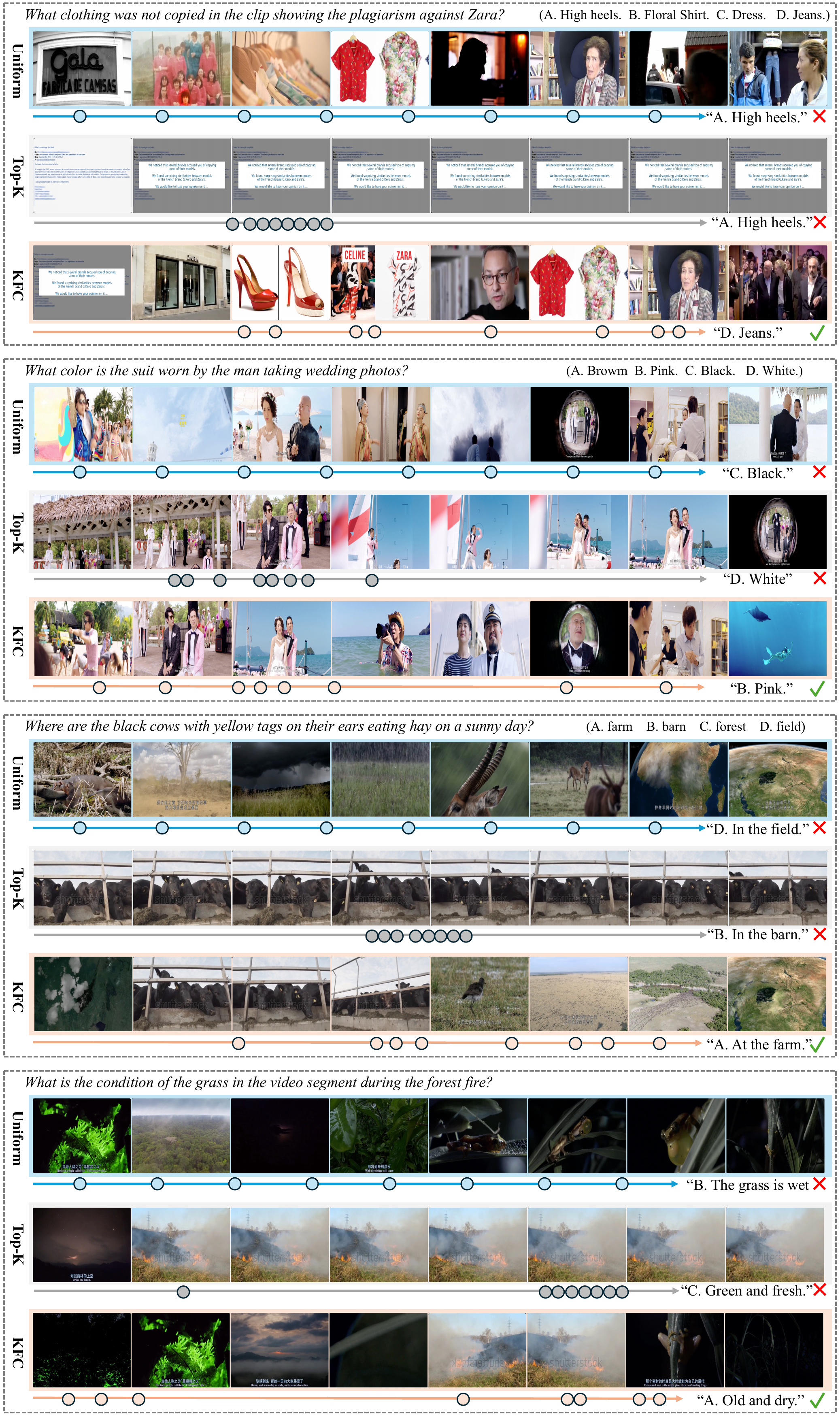}
  \caption{More qualitative examples of keyframe selection using our KFC method, compared with uniform sampling and topK sampling baselines. Zoom in for better visual details.}
  \label{fig:sup_kfc}
\end{figure}

\begin{figure}[ht]
  \centering
  % \fbox{\rule[-.5cm]{0cm}{4cm} \rule[-.5cm]{4cm}{0cm}}
  \includegraphics[width=1.0\textwidth]{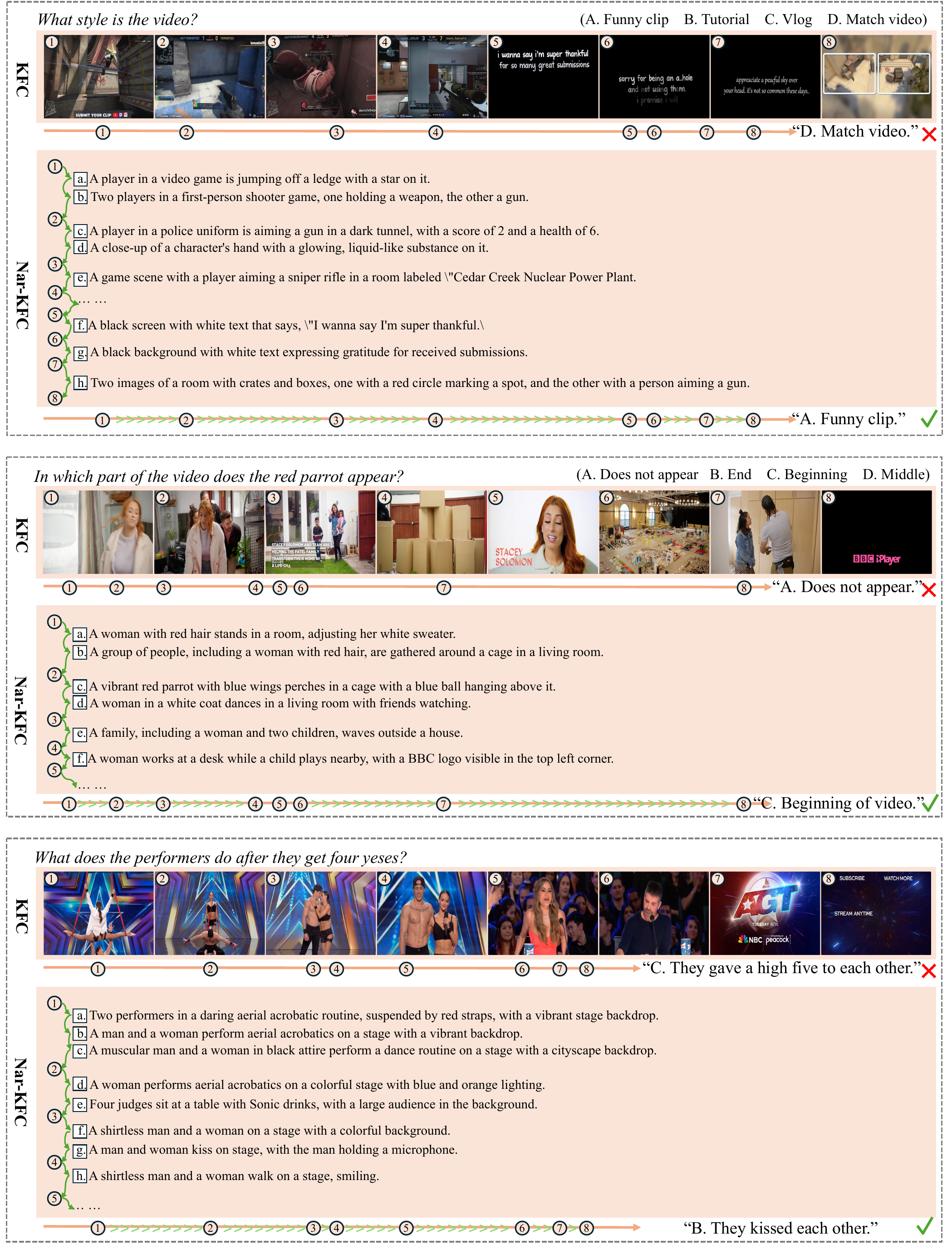}
  \caption{More qualitative examples of our threading keyframe methods Nar-KFC. Zoom in for details.}
  \label{fig:sup_nar_kfc}
\end{figure}

We present additional qualitative examples of our keyframe selection method (KFC) in Fig.~\ref{fig:sup_kfc}, and of the narrating keyframe method (Nar-KFC) in Fig.~\ref{fig:sup_nar_kfc}. 
Note that the frames leading to incorrect predictions in Fig.~\ref{fig:sup_nar_kfc} can be regarded as failure cases of KFC.

\end{document}